\documentclass[12pt, draftcls, onecolumn]{IEEEtran}
\usepackage{amsmath,amsfonts}
\usepackage{algorithmic}
\usepackage{array}
\usepackage[caption=false,font=scriptsize,labelfont=sf,textfont=sf]{subfig}
\usepackage{textcomp}
\usepackage{stfloats}
\usepackage{url}
\usepackage{verbatim}
\usepackage{graphicx}
\def\BibTeX{{\rm B\kern-.05em{\sc i\kern-.025em b}\kern-.08em
    T\kern-.1667em\lower.7ex\hbox{E}\kern-.125emX}}
\usepackage{balance}
\usepackage[font=small,labelfont=bf]{caption}
\usepackage{cite}
\usepackage{enumerate}
\usepackage{enumitem}
\usepackage{bm}
\usepackage{caption}
\usepackage{algorithm}
\usepackage{multirow}
\usepackage{amsthm}

\newtheorem{proposition}{Proposition}
\newtheorem{property}{Property}

\usepackage{xpatch}
\def\*#1{\boldsymbol{\mathbf{#1}}}
\allowdisplaybreaks[3]
\usepackage{titlesec}
\newcommand{\labelsubseccounter}[1]{
    \renewcommand\thesubsection{\Alph{subsection}}
    \addtocounter{subsection}{-1}
    \refstepcounter{subsection}
    \label{#1}
    \renewcommand\thesubsection{\thesection.\Alph{subsection}}
}

\begin{document}
\title{Bayesian Low-rank Matrix Completion \\with Dual-graph Embedding:\\ Prior Analysis and Tuning-free Inference}
\author{Yangge Chen, Lei Cheng, and Yik-Chung Wu
\thanks{Yangge Chen and Yik-Chung Wu are with the Department of Electrical and Electronic Engineering, The University of Hong
Kong, Hong Kong (e-mail:ygchen@eee.hku.hk; ycwu@eee.hku.hk).}
\thanks{Lei Cheng is with the ISEE College, Zhejiang University, Hangzhou, China (email: lei\_cheng@zju.edu.cn).}
}

\markboth{Journal of \LaTeX\ Class Files,~Vol.~18, No.~9, September~2020}%
{How to Use the IEEEtran \LaTeX \ Templates}

\maketitle

\begin{abstract}
Recently, there is a revival of interest in low-rank matrix completion-based unsupervised learning through the lens of dual-graph regularization, which has significantly improved the performance of multidisciplinary machine learning tasks such as recommendation systems, genotype imputation and image inpainting. While the dual-graph regularization contributes a major part of the success, computational costly hyper-parameter tunning is usually involved. To circumvent such a drawback and improve the completion performance, we propose a novel Bayesian learning algorithm that automatically learns the hyper-parameters associated with dual-graph regularization, and at the same time, guarantee the low-rankness of matrix completion. Notably, a novel prior is devised to promote the low-rankness of the matrix and encode the dual-graph information simultaneously, which is more challenging than the single-graph counterpart. A nontrivial conditional conjugacy between the proposed priors and likelihood function is then explored such that an efficient algorithm is derived under variational inference framework. Extensive experiments using synthetic and real-world datasets demonstrate the state-of-the-art performance of the proposed learning algorithm for various data analysis tasks. 
\end{abstract}


\section{Introduction}
\IEEEPARstart{M}{atrix} completion methods, which aim to recover the missing elements from a partially observed matrix, have been studied and applied in various areas, including recommendation system \cite{rennie2005fast,koren2009matrix,bokde2015matrix,rao2017divide}, genotype imputation \cite{chi2013genotype,jiang2016sparrec}, and image processing \cite{hu2012fast,gu2014weighted,xue2017depth,feng2018faster}. The key to the success of matrix completion relies on exploiting the intrinsic low-rank structure of the real-world data matrix $\mathbf{X} \in \mathbb{R}^{m \times n}$. This makes the low-rank matrix completion problem being formulated as:
\begin{align}
\label{mc01}
    &\min_{\mathbf{X} \in \mathbb{R}^{m \times n}} \mathrm{rank}(\mathbf{X}) \nonumber
    \\&s.t.\quad \mathcal{P}_{\Omega}(\mathbf{M}) = \mathcal{P}_{\Omega}(\mathbf{X}),
\end{align}
where $\mathbf{M} \in \mathbb{R}^{m \times n}$ is a partially observed matrix, $\mathbf{\Omega}$ is an indicator matrix with $\mathbf{\Omega}_{ij} = 1$ if $\mathbf{M}_{ij}$ is observed and zero otherwise, and $\mathcal{P}_{\Omega}(\cdot)$ is a projection operator which retains the entries of matrix in the positions with $\mathbf{\Omega}_{ij} = 1$. 

As rank minimization is NP-hard and the hard constraint in (\ref{mc01}) is not robust to observation noise \cite{candes2010matrix,ma2011fixed}, problem (\ref{mc01}) is normally relaxed to:
\begin{equation}
    \min_{\*X} \quad\lambda\|\mathbf{X}\|_* + ||\mathcal{P}_{\Omega}(\mathbf{M}-\mathbf{X})||_{F}^2,
    \label{mc03}
\end{equation}
where $||\cdot||_*$ denotes the nuclear norm, the tightest convex relaxation of $\rm rank(\cdot)$ \cite{srebro2005maximum}, and $\lambda$ is a parameter specifying the relative importance of the two terms of (\ref{mc03}). 

To address the high complexity issue brought by nuclear norm minimization \cite{zheng2015convergent,shah2016biconvex}, factorized model $\mathbf{X} = \mathbf{U}\mathbf{V}^T$ has been proposed \cite{candes2009exact}, with the corresponding completion problem formulated as 
\begin{equation}
    \min_{\mathbf{U},\mathbf{V}} \quad \frac{\lambda_u}{2}||\mathbf{U}||_F^2 + \frac{\lambda_v}{2}||\mathbf{V}||_F^2 + ||\mathcal{P}_{\Omega}(\mathbf{M}-\mathbf{U}\mathbf{V}^T)||_{F}^2,
    \label{mc03_fac}
\end{equation}
where the number of columns $k$ in $\mathbf{U}$ and $\mathbf{V}$ is much smaller than $m$ or $n$. Specifically, problem (\ref{mc03_fac}) replaces the nuclear norm in (\ref{mc03}) with the Frobenius norms of factor matrices, and a small value of $k$ implies the low-rankness of $\mathbf{U}\mathbf{V}^T$. 
Compared to the model (\ref{mc03}), factorized model (\ref{mc03_fac}) provides explicit latent feature spaces, which enable additional interpretation of the unsupervised learning. For example, in the famous example of Netflix movie recommendation, the user-movie rating matrix $\mathbf{M}$ is factorized into user-feature matrix $\mathbf{U}$ and movie-feature matrix $\mathbf{V}$, where $k$ is the number of features. 
This means that the movies are grouped into $k$ major categories despite the larger number of original movie genres. It is the same for the user preferences clustering, which contains different grouping from given user profiles. Due to the scalability and interpretability, the factorized model has been central to the research of collaborative filtering and recommendation system for over two decades \cite{mehta2017review}.

In practical applications, the observation matrix $\*M$ is usually very sparse.  For example, in movie rating recommendation, most users have only a few rating histories. Similar challenges have also been encountered in other unsupervised learning tasks including genotype imputation \cite{chi2013genotype,jiang2016sparrec}, and image processing \cite{hu2012fast,gu2014weighted,xue2017depth,feng2018faster}. To deal with the difficulty caused by the sparsity of nowadays data, recent matrix completion methods tend to incorporate the domain knowledge/side information into the algorithm design, among which the graph regularization-based methods have received the most attention. Below, we review related low-rank matrix completion works with graph regularization.



\subsection{Related Works: An Optimization Way}
The matrix completion problem (\ref{mc03}) and (\ref{mc03_fac}) mainly focus on exploiting the low-rank property of the data matrix $\mathbf{X}$. If graph information about $\mathbf{X}$ is available, it can be incorporated to improve its recovery performance. In particular, assume the rows of $\mathbf{X}$ represent a set of vertices $\mathcal{V}$ of an undirected graph $(\mathcal{V},\mathcal{E})$, where $\mathcal{E} \subseteq \{(i,j)|i,j \in \mathcal{V}\}$ is the set of edges. The existence of an edge implies that connected vertices are supposed to be close and similar. To exploit these pairwise similarities between the vertices (rows of $\mathbf{X}$), a graph regularization term is defined as:
\begin{equation}
    \sum_{i,j} \mathbf{A}_{ij}||\mathbf{x}_i - \mathbf{x}_j||^2_2 = {\rm tr}(\mathbf{X}^T \mathbf{L}_r \mathbf{X}),
\label{Lreg}
\end{equation}
where $\mathbf{x}_i$ is the $i^{th}$ row of $\mathbf{X}$, $\mathbf{L}_r = \mathbf{D} - \mathbf{A}$ is the graph Laplacian matrix \cite{chung1997spectral}, $\mathbf{A}$ is the adjacency matrix with $\mathbf{A}_{ij}$ being one if $(i,j)\in \mathcal{E}$ and zero otherwise, and $\mathbf{D}$ is a diagonal matrix with $\mathbf{D}_{ii} = \sum_j \mathbf{A}_{ij}$. Adding this term to the objective function of (\ref{mc03}) leads to the graph-regularized matrix completion problem:
\begin{equation}
\min_{\mathbf{X}} \quad \|\mathbf{X}\|_* +  \tau||\mathcal{P}_{\Omega}(\mathbf{M}-\mathbf{X})||_{F}^2 + \gamma {\rm tr}(\mathbf{X}^T \mathbf{L}_r \mathbf{X}),
\label{mc04}
\end{equation}
where the solution will be steered to have the property $\mathbf{x}_i\approx \mathbf{x}_j$ if $(i,j) \in \mathcal{E}$ \cite{shang2012graph,zhang2012low,jiang2013graph,shahid2015robust}. 

In real-world applications, we often have access to both row and column graphical structures. It is then natural to extend the problem (\ref{mc04}) to a dual-graph regularized problem \cite{kalofolias2014matrix,shahid2016fast}:
\begin{align}
\label{mc05}
\min_{\mathbf{X}} \quad & \|\mathbf{X}\|_* + \tau||\mathcal{P}_{\Omega}(\mathbf{M}-\mathbf{X})||_{F}^2 + \gamma_r {\rm tr}(\mathbf{X}^T \mathbf{L}_r \mathbf{X}) + \gamma_c {\rm tr}(\mathbf{X} \mathbf{L}_c \mathbf{X}^{T}),
\end{align}
where $\mathbf{L}_c$ and $\mathbf{L}_r$ are the Laplacian matrix of the given column graph and row graph of $\mathbf{X}$ respectively. Problem (\ref{mc05}), which incorporates dual-graph information, involves two more regularization parameters than (\ref{mc03}). Once these regularization parameters $\{\tau, \gamma_r, \gamma_c\}$ are fixed, problem (\ref{mc05}) can be solved by Alternating Direction Method of Multipliers (ADMM) \cite{kalofolias2014matrix}. 

On the other hand, if the graph regularizations are applied to the factorized model (\ref{mc03_fac}), the corresponding problem becomes
\begin{align}
\min_{\mathbf{U},\mathbf{V}} \quad & \frac{\gamma_u}{2}||\mathbf{U}||_F^2 + \frac{\gamma_v}{2}||\mathbf{V}||_F^2 + \frac{\tau}{2} ||\mathcal{P}_{\Omega}(\mathbf{U}\mathbf{V}^T - \mathbf{M})||_{F}^{2}+ \frac{1}{2} \{{\rm tr}(\mathbf{U}^T\mathbf{L}_{r}\mathbf{U}) + {\rm tr}(\mathbf{V}^{T}\mathbf{L}_{c}\mathbf{V})\}.
\label{opt-fac}
\end{align}
The two factor matrices $\mathbf{U}^{m \times k}$ and $\mathbf{V}^{n \times k}$ capture the row and column graph information of $\*X$ respectively. 
Since the minimization problem in (\ref{opt-fac}) is block multi-convex with respect to $\mathbf{U}$ or $\mathbf{V}$ \cite{chen2016direct}, an efficient alternating optimization algorithm, which solves $\mathbf{U}$ while $\mathbf{V}$ is fixed and solves $\mathbf{V}$ while $\mathbf{U}$ is fixed, can be applied \cite{rao2015collaborative}.

Although problem (\ref{mc05}) and (\ref{opt-fac}) achieve promising recovery performance when the regularization parameters $\{\gamma_u, \gamma_v, \tau\}$ are carefully set, tunning these hyperparameters is not an easy task. It could be very time-consuming, and tuning must be redone for each dataset. More importantly, parameter tuning may not even be feasible in real-world applications where the ground-truth of $\mathbf{X}$ is unavailable.
For the factorized model (\ref{opt-fac}), an additional latent rank $k$ of $\mathbf{U}^{m \times k}$ and $\mathbf{V}^{n \times k}$ is required. A proper rank should follow the low-rankness of data meanwhile be sufficiently large to avoid the underfitting of data. The selection of a proper $k$ is always a challenge in the factorized models (especially with the missing data): the computation of the optimal $k$ is NP-hard, and a suboptimal $k$ would significantly affect the recovery performance \cite{shi2017rank}.

\begin{figure}[t]
\setlength{\belowcaptionskip}{-0.5cm}
\subfloat[\scriptsize{RMSE vs. graph kernel parameter}\label{sen_KPMF}]{%
        \includegraphics[width=0.4\linewidth]{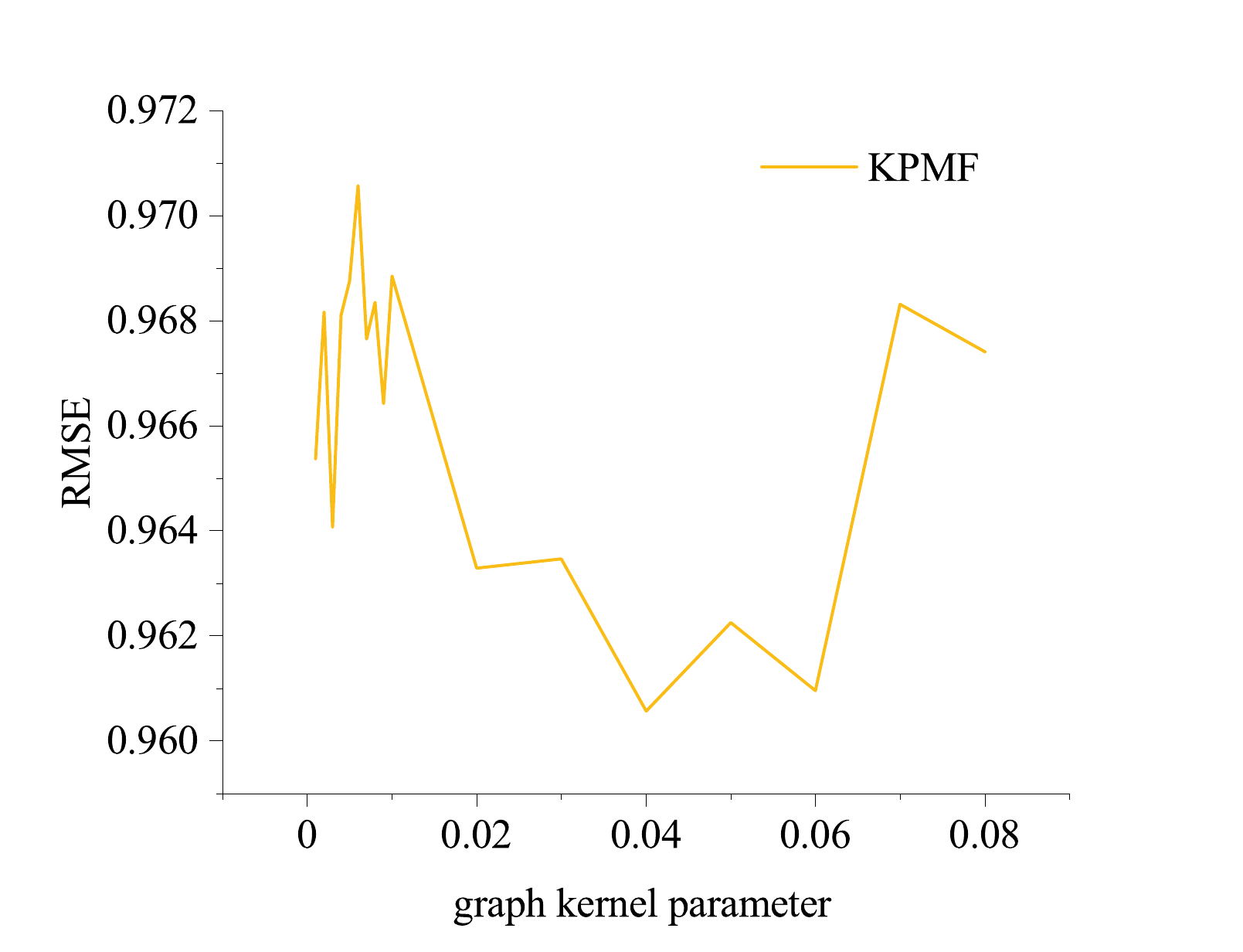}}
\hfill
\subfloat[\scriptsize{RMSE vs. graph regularization parameters}\label{sen_GRALSGPMF}]{%
        \includegraphics[width=0.4\linewidth]{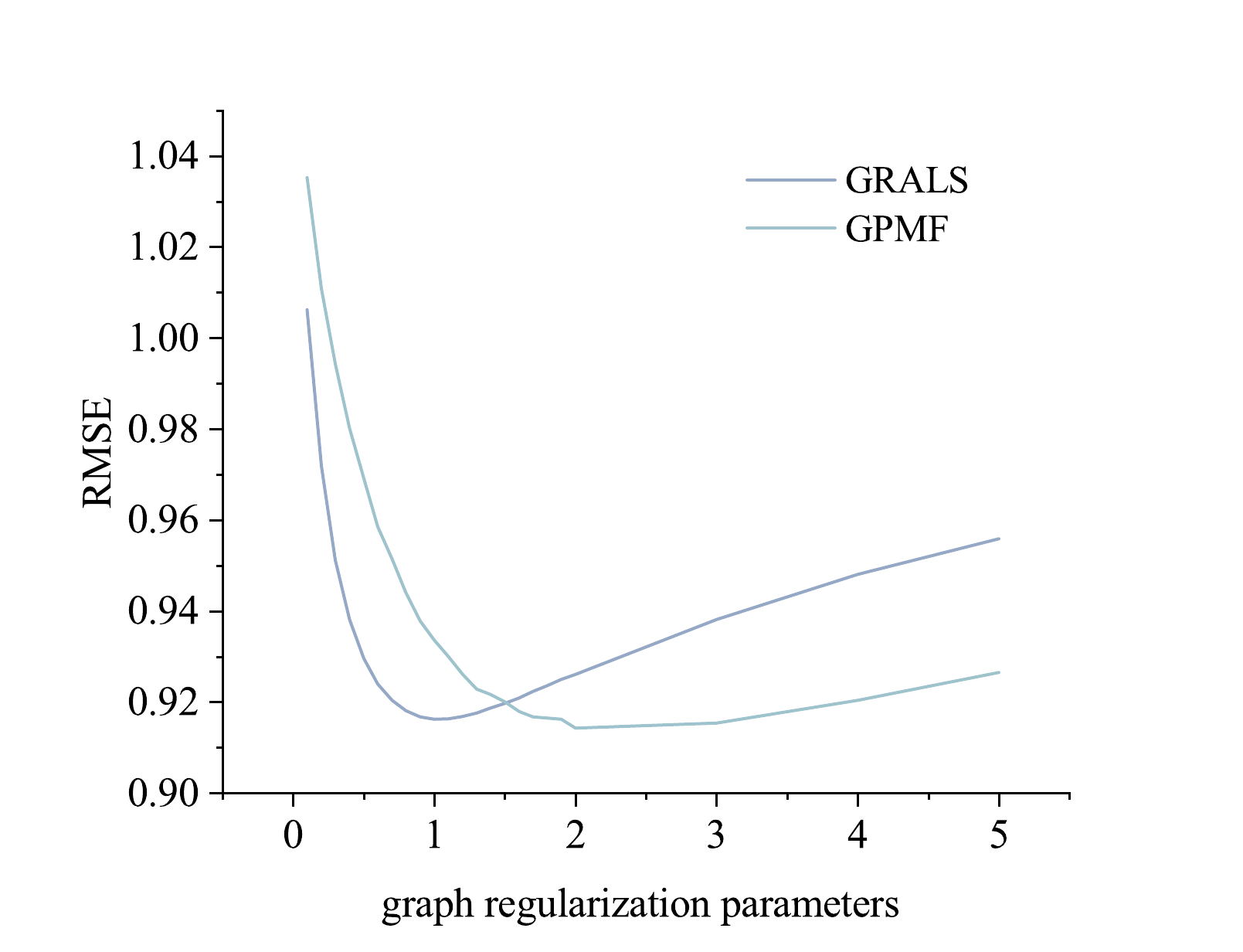}}
\\
\subfloat[\scriptsize{graph regularization parameters $= 0.05$}\label{sen_GRALS}]{%
        \includegraphics[width=0.20\linewidth]{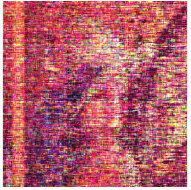}}
\hfill
\subfloat[\scriptsize{graph regularization parameters $= 0.1$}\label{sen_GRALS2}]{%
        \includegraphics[width=0.20\linewidth]{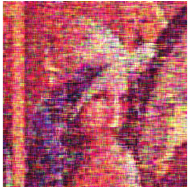}}
\hfill
\subfloat[\scriptsize{graph regularization parameters $= 0.5$}\label{sen_GRALS3}]{%
        \includegraphics[width=0.20\linewidth]{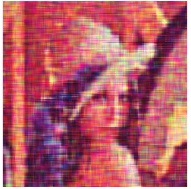}}
\hfill
\caption{Parameter sensitivities for three state-of-the-art graph-regularized matrix completion methods: KPMF \cite{zhou2012kernelized}, GRALS \cite{rao2015collaborative} and GPMF \cite{strahl2020scalable}. \protect\subref{sen_KPMF} and \protect\subref{sen_GRALSGPMF} are on the popular movie-recommendation datasets Movielens $100k$. \protect\subref{sen_GRALS} - \protect\subref{sen_GRALS3} show the completed images of GRALS under different graph regularization parameters $\gamma_r, \gamma_c$ with latent rank and other regularization parameters fixed.}
\label{sensitivity_intro} 
\end{figure}

To show the importance of choosing the ``proper'' regularization parameters, Fig. \ref{sensitivity_intro} shows the completion results in different learning tasks by existing algorithms. It is obvious that the missing data recovery performances are highly sensitive to the choices of hyperparameters. Due to this reason, all prior works require exhaustive tuning of regularization parameters to obtain satisfactory performance. 

\subsection{Related Works: A Bayesian Way}
It is noteworthy that the regularization terms in problem \eqref{opt-fac} can be interpreted in terms of a Bayesian prior model. 
In particular, the Frobenius norm of $\mathbf{U}$ and graph regularized term on $\mathbf{U}$ can be combined and interpreted as a zero-mean Gaussian prior with precision matrix being $(\mathbf{L}_r + \gamma_u \mathbf{I})$. Similarly, $\*V$ can also be interpreted as having a zero-mean Gaussian prior with precision matrix $(\*L_c + \gamma_v \*I)$. Furthermore, with the third term of \eqref{opt-fac} being interpreted as the likelihood function, the optimization problem \eqref{opt-fac} can be interpreted as finding the maximum a posterior (MAP) solution of a Bayesian model \cite{zhou2012kernelized,strahl2020scalable}:
\begin{equation}
    \max_{\*U,\*V} \quad p(\*M|\*U,\*V,\tau)p(\*U)p(\*V),
\end{equation}
where
\begin{align}
    \label{opt_U} p(\*U) &= \mathcal{N}(\*U|\*0,\*I,(\*L_r + \gamma_u \*I)^{-1}),\\
    \label{opt_V} p(\*V) &= \mathcal{N}(\*V|\*0,\*I,(\*L_c + \gamma_v \*I)^{-1}),\\
    p(\*M|\*U,\*V,\tau) &= \prod_{(i,j)\in\*\Omega}\mathcal{N}(\*M_{ij}|\*U_{i:}\*V_{j:}^T,\tau^{-1}).
\end{align}
However, interpreting \eqref{opt-fac} using a Bayesian perspective alone does not mitigate the problem of choosing regularization parameters ($\{\gamma_u, \gamma_v, \tau\}$ and matrix rank $k$). The problem persists since there is no Bayesian modeling of the regularization parameters.

To overcome the difficulty in hyper-parameter tuning, fully Bayesian learning, in which prior distribution is also imposed on the hyper-parameters, is of great importance in the research of matrix completion.
In particular, in basic matrix completion without graph information \cite{babacan2012sparse}, $\tau$ is interpreted as the noise precision and a gamma prior distribution is imposed, so that $\tau$ is learned together with $\*U$ and $\*V$. Furthermore, instead of separate graph regularization parameters $\gamma_u$ and $\gamma_v$, a common $\gamma_i$ is shared by the $i^{th}$ column of $\*U$ and $\*V$ to impose sparsity in the factorized model. Similar to $\tau$, each $\gamma_i$ will be modeled as a Gamma distributed hyperparameter. Subsequently, fully Bayesian learning has been applied to various matrix/tensor completion works \cite{chen2021bayesian,xu2021overfitting,li2020bayesian,zhao2015bayesian,zhao2015bayesianrobust,paliwal2021traffic,cheng2016probabilistic} with different side information. These methods demonstrate that hyperparameters tuning can be avoided and can yield the same or even better data recovery accuracy than the optimization-based approaches with costly tuning.

Given the interpretability of factorized models and dual-graph regularization in optimization problem \eqref{opt-fac}, an obvious next step is to study an automatic parameter tuning version of \eqref{opt-fac} by using fully Bayesian learning. Recently, \cite{yang2018fast} proposed a fully Bayesian algorithm for the matrix completion problem with a single row graph as the scale matrix of the Gaussian-Inverse Wishart hierarchical prior model. While this is an inspiring and pioneering work, it cannot handle factorized models and embed dual-graph information.

The major challenge in developing a fully Bayesian dual-graph factorization model is the lack of suitable prior distributions for the factor matrices. In particular, in the factorization model $\*X = \*U\*V^T$, the low-rankness of $\*X$ is reflected in the small number of columns in $\*U$ and $\*V$.  Furthermore, the dual-graph information of $\*X$ is embedded in the row correlation of $\*U$ and $\*V$ respectively. Thus, to develop the fully Bayesian dual-graph factorization model, a factor matrix prior with simultaneous incorporation of graph information \textit{among its rows} and sparsity \textit{across its columns} is needed. In addition, if we want to develop efficient inference algorithm, the likelihood function and the prior need to satisfy certain conditional conjugacy \cite{beal2003variational,bishop2006pattern,winn2005variational}. While identifying conjugacy has been done by many previous works, accomplishing it under missing data and non-diagonal row correlation matrix turns out to be not as straightforward as it seems.

\subsection{Contributions}
\begin{figure}[t]
\setlength{\belowcaptionskip}{-0.5cm} 
    \centering
    \subfloat[Movielens $100k$\label{sen_ours_sota}]{%
        \includegraphics[width=0.5\linewidth]{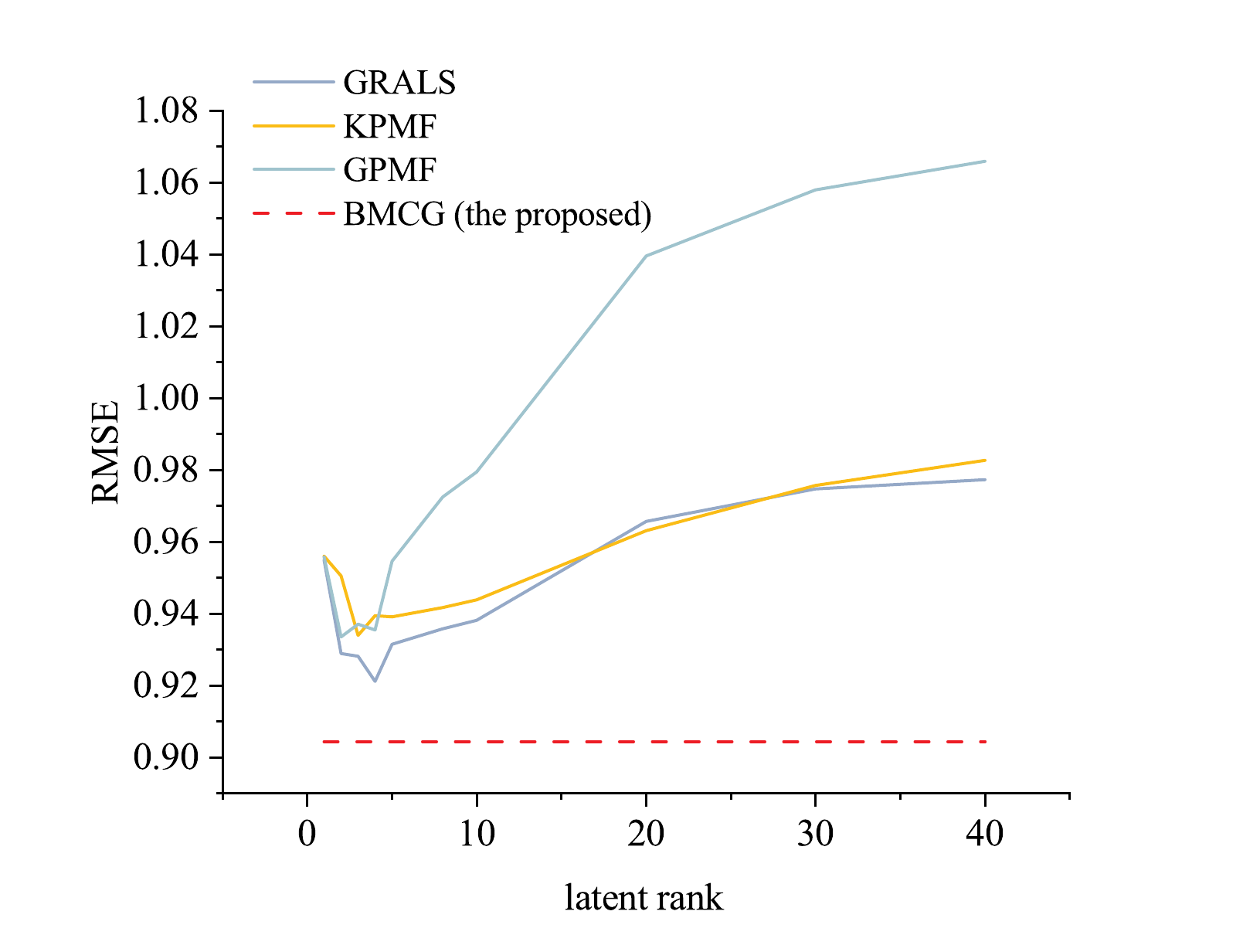}}
\hfill
\subfloat[Image recovery performances of cross-masked images\label{stripimg}]{%
\includegraphics[width=1\linewidth]{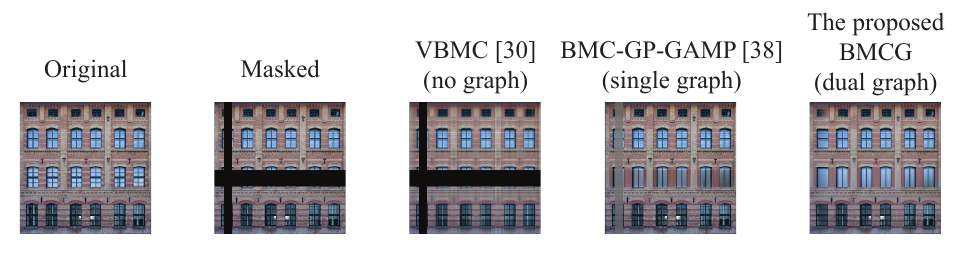}}
\hfill
    \caption{Performance comparison in movie recommendation and image recovery}
    \label{comparisonintro}
\end{figure}
In this paper, we develop a novel Bayesian matrix completion model with simultaneous graph embedding and low-rankness (termed as BMCG). Our major contributions are summarized as follow:
\begin{itemize}[leftmargin=*]
\item \textbf{Design and analysis of a dual-graph and low-rankness embedding prior}: We first establish a new prior model for the factor matrix. It is proved with theoretical analysis that the proposed prior is effective in expressing row-graph information while imposing column-wise sparsity. Applying the novel priors to the factor matrices in a matrix factorization model results in simultaneous dual-graph regularization and matrix low-rankness.
\item \textbf{Conditional conjugacy and column-wise mean-field factorization}: Due to the missing data and non-diagonal row graph Laplacian matrix, the proposed prior of $\*U$ and $\*V$ are not directly conjugate to the likelihood function. We mathematically confirm the conditional conjugacy between the newly designed prior and the likelihood function when the prior of $\*U$ and $\*V$ are viewed in terms of their columns, which is the stepping stone towards efficient Bayesian inference. The resulting column-wise mean-field factorization strikes a good balance between reducing information loss during inference and improving algorithmic efficiency.
\item \textbf{Effectiveness in multiple real-life datasets and applications}: Extensive experiments show that the proposed method exhibits superior performances in missing data recovery in synthetic data and multidisciplinary real-life applications, including movie recommendation systems, gene inputting, and image inpainting in a tuning-free fashion. As a sneak peek, performance comparison with three state-of-the-art dual-graph regularized matrix completion algorithms is shown in Fig. \ref{comparisonintro}\subref{sen_ours_sota} and the comparisons with two previous fully Bayesian algorithms (without/with single graph information) are shown in Fig. \ref{comparisonintro}\subref{stripimg}.
\end{itemize}

The remainder of this paper is organized as follows. Section \ref{sec2} analyzes the proposed random matrix distributions and shows how it simultaneously embeds the dual-graph information and the low-rank structure. Furthermore, the conditional conjugacy between the proposed priors and likelihood function is explored in this section. Section \ref{sec3} derives the column-wise variational inference algorithm and presents the insights into the updating equations. Numerical results are demonstrated in Section \ref{sec4}. Finally, the conclusions are drawn in Section \ref{sec_con}.

\section{Simultaneous Dual-graph Information Embedding and Low-rankness}
\label{sec2}
Before moving to the design of prior for Bayesian matrix completion, we first propose a prior for a random matrix that encodes the graph information and the sparsity structure simultaneously. Then, by noticing that the probabilistic modeling of low-rankness can be recast as the modeling of column-wise sparsity in the factor matrices $\*U$ and $\*V$, we use the proposed random matrix prior to build a novel prior for the Bayesian matrix completion problem, with appealing features of low-rankness promoting and dual-graph information embedding. 

\subsection{New Prior for a Random Matrix}
\begin{figure}
\centering
\subfloat[\scriptsize{Graph-embedding and sparsity promoting prior}\label{priorL}]{%
        \includegraphics[width=0.45\linewidth]{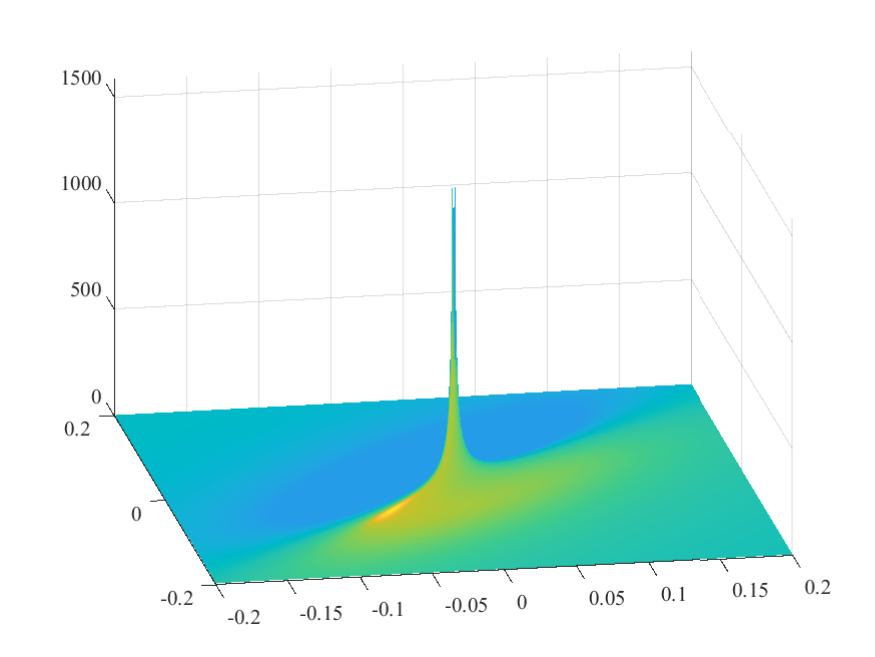}}
\hfill
\subfloat[\scriptsize{Bivariate student-t prior}\label{priorI}]{%
        \includegraphics[width=0.45\linewidth]{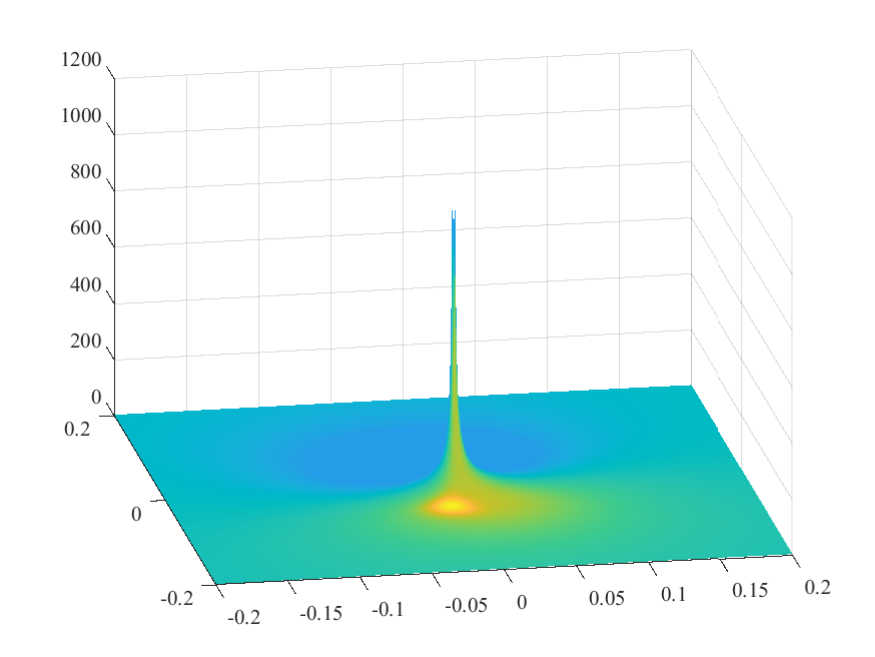}}
\caption{Demonstration of the marginal prior distribution of $\*w \in \mathbb{R}^{2\times 1}$ in (\ref{proposition-1}) when hyperparameters tend to zero. Left: with $\*L=[1,-0.8;-0.8,1]$, resulting in a sparsity-promoting (highly peaked at zero and having heavy tails) and graph-embedded ($w_1\approx w_2$) distribution; Right: a standard student-t distribution (i.e., by setting $\*L=\*I$ in (\ref{proposition-1})), which is only sparsity-promoting (peaked at zero and having heavy tails).}
\label{compareWishart} 
\end{figure}

For a random matrix $\*W \in \mathbb{R}^{m\times n}$, we propose a new matrix normal prior by setting the graph Laplacian matrix $\*L$ as the row precision matrix (the inverse of row covariance matrix \cite{bernardo2009bayesian}), and $\*\Lambda = {\rm diag}(\lambda_1,...,\lambda_n)$ as the column precision matrix \cite{gupta2018matrix}, that is,
\begin{equation}
    p(\*W|\*\lambda) = \mathcal{N}(\*W|\*0,\*\Lambda^{-1},\*L^{-1}),
    \label{xlambda-x}
\end{equation}
with $\*\lambda = (\lambda_1,...,\lambda_n)$ modeled by:
\begin{equation}
    p(\*\lambda) = \prod_{r=1}^n Ga(\lambda_r|c_0^r,d_0^r).
    \label{xlambda-l}
\end{equation}
This hierarchical construction surprisingly endues the marginalized prior $p(\*W)$ with two important functionalities: graph information embedding and sparsity promoting. To see this, we present the following proposition. 
\begin{proposition}
The marginalized prior distribution of $\*W$ is
\begin{equation}
    p(\*W|\bm{c_0},\bm{d_0}) = \prod_{r=1}^n
    \frac{(d_0^r)^{c_0^r}\sqrt{\*L}\Gamma(c_0^r+\frac{1}{2})}{\sqrt{(2\pi)^{m}}\Gamma(c_0^r)}(d_0^r+\frac{\*w_r^T \*L \*w_r}{2})^{-(c_0^r+\frac{1}{2})},
\label{proposition-1}
\end{equation}
where $\bm{c_0} = (c_0^1,...c_0^n)$ and $\bm{d_0} = (d_0^1,...d_0^n)$. The prior (\ref{proposition-1}) is column-wise sparsity-promoting and also embeds the graph information. 
\label{proposition1}
\end{proposition}
Equation (\ref{proposition-1}) can be obtained by multiplying \eqref{xlambda-x} and \eqref{xlambda-l} and then integrating out $\{\lambda_1,...,\lambda_n\}$. As can be seen from Proposition \ref{proposition1}, the prior distribution of each column of $\*W$ would remain a multivariate t-distribution with scale matrix $\*L^{-1}$. For the $r^{th}$ column of $\*W$, i.e., $\*w_r$, when the hyperparameters $\{c_0^r,d_0^r\}$ tend to zero, the prior distribution is $p(\*w_r) \propto (\*w_r^T \*L\*w_r)^{-\frac{1}{2}}$, which peaks at zero and has a heavy tail. In this way, the resulting $\*w_r$ is most likely to be a zero vector while also allows to take some large values, thus enforcing column-wise sparsity. When $\*w_r$ is not a zero-vector, it is regulated by the underlying graph structure. A demonstration of the simultaneous embedding of sparsity and graph information is shown in Fig. \ref{compareWishart}\subref{priorL}, and as a comparison, the standard bivariate student-t distribution obtained by setting $\*L=\*I$ in \eqref{proposition-1} is shown in Fig. \ref{compareWishart}\subref{priorI}.

\begin{figure}[t]
        \centering
        \setlength{\belowcaptionskip}{-0.5cm}
        \includegraphics[width=0.3\linewidth]{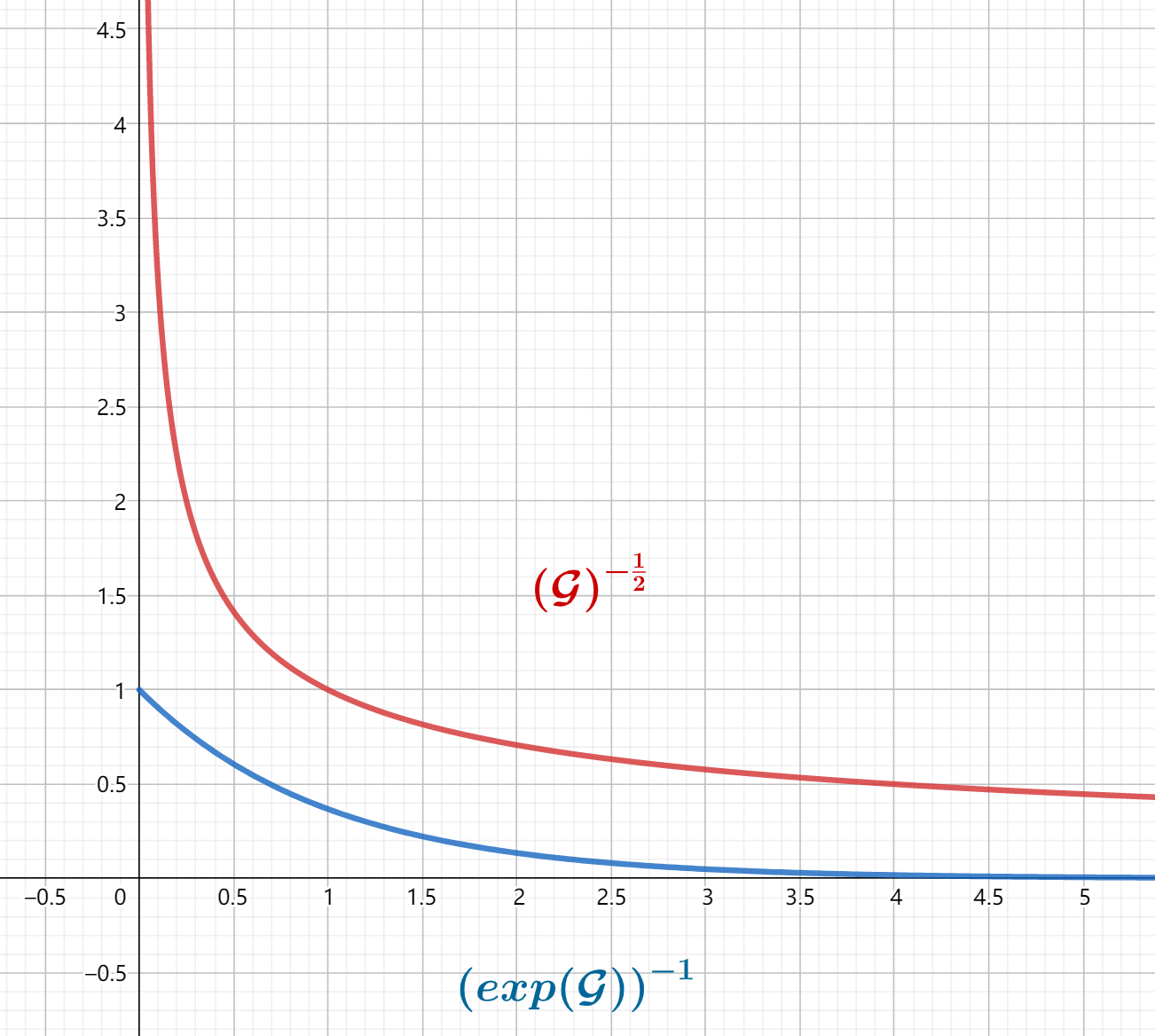}
        \caption{$p(\*w)$ versus graph regularization $\mathcal{G}$ with the proposed prior (red) and prior in \cite{yang2018fast} (blue).}
        \label{fig:compareWishart}
\end{figure}

Recently, another prior distribution for Bayesian matrix completion problem was proposed as $p(\*W|\*\Sigma) = \mathcal{N}(\*0,\*\Sigma^{-1},\*I)$ with $\*\Sigma$ obeying Wishart distribution, and the scale matrix of the Wishart distribution is set as the graph Laplacian matrix $\*L$ \cite{yang2018fast}. However, the graph information embedding of the proposed model \eqref{xlambda-x} is more effective since the Laplacian matrix $\*L$ in the model of \cite{yang2018fast} goes through the Wishart distribution before regularizing $\*W$, while the proposed model directly applies regularization to $\*W$.
It can be shown that the marginal prior of $\*W$ in \cite{yang2018fast} is $p(\*W) \propto \prod_r ({\rm exp}(\*w_r^T \*L \*w_r))^{-1}$. To compare this prior with $p(\*W) \propto \prod_r (\*w_r^T \*L \*w_r)^{-\frac{1}{2}}$, we let $\mathcal{G} = \*w_r^T \*L\*w_r$ and plot $({\rm exp}(\mathcal{G}))^{-1}$ and $(\mathcal{G})^{-\frac{1}{2}}$ in Fig. \ref{fig:compareWishart}.

Since $\*L=(\sum \*A_{1,j} ... \sum \*A_{n,j}) -\*A$, $\*L$ is a matrix with off-diagonal element being negative, with their magnitudes indicating how strong the graph structure is. In the extreme case when there is no graph information, $\*L$ is a diagonal matrix. Therefore, for a given $\*w_r$, $\mathcal{G}$ would take large values if graph structure is weak, and small values if graph structure is strong. It can be seen from Fig. \ref{fig:compareWishart} that, on one hand, when graph structure is weak ($\*L$ is close to $\*I$, and $\mathcal{G}$ tends to be large), both priors have relatively flat responses to the graph regularization, meaning that both models barely use the graph information. On the other hand, when the graph structure is strong and informative ($\mathcal{G}$ close to zero), the proposed prior shows a more discriminative response than the one proposed in \cite{yang2018fast}. Due to the more direct usage of the graph Laplacian matrix $\*L$, the proposed prior model leverages graph information \textit{more effectively}.

Besides the difference in the marginal prior distribution, the model in \cite{yang2018fast} can only make use of row graph information, whereas the proposed model in this paper can incorporate \textit{dual-graph information}, as shown in the next subsection.

\subsection{Dual-graph and Low-rankness Embedding}
Given the appealing features of the proposed random matrix prior \eqref{xlambda-x}-\eqref{xlambda-l}, we now consider to use it as the cornerstone towards the prior design of Bayesian matrix completion. Note that the latent matrix $\*X$ is factorized as
\begin{equation}
    \*X = \sum_{r=1}^k \*u_r \*v_r^T,
\label{Xfactor}
\end{equation}
where $\*u_r$ is the $r^{th}$ column of $\*U$ and $\*v_r$ is the $r^{th}$ column of $\*V$. 
Since the matrix rank is defined as the number of independent rank-1 component $\*u_r\*v_r^T$, if most column vectors $\*u_r$ and $\*v_r$ are zero, the resulting matrix $\*X$ is low-rank. This suggests that the modeling of low-rankness is equivalent to the modeling of column-wise sparsity for $\*U$ and $\*V$.

To further incorporate the dual-graph information, assume that we are provided with a graph $(\mathcal{V}_r,\mathcal{E}_r)$ which encodes the pairwise relationship between the rows of $\*U$ and $(\mathcal{V}_c,\mathcal{E}_c)$ which encodes the pairwise relationship between the rows of $\*V$. 
Based on the discussions in the last subsection, we place prior distributions over each latent matrices $\*U$ and $\*V$ with a shared hyperparameter set $\bm{\lambda} = \{\lambda_1,...,\lambda_k\}$ and graph Laplacian matrices $\*L_r$ and $\*L_c$ as follows:
\begin{align}
    \label{U} p(\*U|\bm{\lambda})& = \mathcal{N}(\*U|\*0,\*\Lambda^{-1},\*L_r^{-1}),\\
    \label{V} p(\*V|\bm{\lambda})& = \mathcal{N}(\*V|\*0,\*\Lambda^{-1},\*L_c^{-1}),\\
    \label{lambda} p(\bm{\lambda})& = \prod_{r=1}^k Ga(\lambda_r|c_0^r,d_0^r),
\end{align}
where $\*\Lambda = {\rm diag}(\bm{\lambda})$. Applying Proposition \ref{proposition1} to \eqref{U}-\eqref{lambda}, we obtain the following proposition.

\begin{proposition}
The marginalized priors for $\*U$ and $\*V$ are given by \eqref{proposition-1} with $\*W$ replaced by $\*U$ and $\*V$ respectively.
Furthermore, when $\bm{c_0} = (c_0^1,...c_0^k)$ and $\bm{d_0}= (d_0^1,...d_0^k)$ go to zeros, they become
\begin{align}
    p(\*U) \propto \prod_{i=1}^k (\*u_i^T \*L_r \*u_i)^{-\frac{1}{2}},\quad
    p(\*V) \propto \prod_{j=1}^k (\*v_j^T \*L_c \*v_j)^{-\frac{1}{2}}.
\end{align}
\label{proposition2}
\end{proposition}
Therefore, the priors of $\*U$ and $\*V$ are highly peaked at zero and having heavy tails. This will drive most of the columns of $\*U$ and $\*V$ to zero, thus encouraging the column-wise sparsity of both factor matrices and naturally a small rank of $\*X = \*U\*V^T$.
To see how the sparsity is induced while maintaining dual-graph information, we provide the following proposition.  
\begin{proposition}
The logarithm of prior is given:
\begin{equation}
    \ln p(\*U,\*V|\bm{\lambda})
    \propto -\frac{1}{2}\sum_{i=1}^k\lambda_i (\*u_i^T \*L_r \*u_i + \*v_i^T \*L_c \*v_i)
\label{logu}
\end{equation}
\begin{proof}
As $\ln p(\*U,\*V|\bm{\lambda}) \propto \ln p(\*V|\bm{\lambda})p(\*U|\bm{\lambda})$, equation \eqref{logu} can be obtained by multiplying \eqref{U} and \eqref{V} together.
\end{proof}
\label{proposition3}
\end{proposition}
As can be seen from Proposition \ref{proposition3}, to maximize the prior distribution, when the shared hyperparameter $\lambda_i$ takes a large value, the $i^{th}$ column of $\*U$ and $\*V$ will tend to be zero columns. For the remaining columns, where the $\lambda_i$s have smaller values, the dual-graph information $\*L_c$ and $\*L_r$ could be incorporated as regularization. However, the regularization in \eqref{logu} is different from the regularization in \eqref{opt-fac} with Bayesian interpretation in \eqref{opt_U} and \eqref{opt_V}, as \eqref{logu} has a separate regularization parameter for each column of $\*U$ and $\*V$, while \eqref{opt-fac} applies a single regularization parameter to the whole matrix $\*U$ and $\*V$. In fact, \eqref{logu} makes more sense since it not only enhances model flexibility but also provides a \textit{trade-off} between model fitting and graph regularization for each column, which will be elaborated in the next section.

Propositions \ref{proposition2} and \ref{proposition3} thus show that the prior model \eqref{U}-\eqref{lambda} enable matrix $\*M$ to encode dual-graph information and let factor matrices $\*U$ and $\*V$ achieve column-wise sparsity simultaneously. 

\subsection{Proposed Probabilistic Model}
 With the prior model established in \eqref{U}-\eqref{lambda}, we need the likelihood function to complete the model. In particular, the observed matrix $\*M$ is expressed as
\begin{equation}
    \mathcal{P}_{\Omega}(\*M) = \mathcal{P}_{\Omega}(\*U\*V^T) + \mathcal{P}_{\Omega}(\*E)
\label{m=uve}
\end{equation}
where $\*\Omega$ is the indicator matrix, $\*U \in \mathbb{R}^{m \times k}$, $\*V \in \mathbb{R}^{n \times k}$ are the factor matrices with $k$ being the latent rank and $\*E$ denotes the additive noise.
The entries of $\*E$ are assumed to be independently and identically distributed (i.i.d.) Gaussian random variables with zero mean and variance $\tau^{-1}$, which gives rise to the likelihood model:
\begin{equation}
    p(\*M|\*U,\*V,\tau) = \prod_{(i,j)\in \*\Omega} \mathcal{N}(\*M_{ij}|(\sum_{r=1}^{k}\*u_r \*v^T_r)_{ij},\tau^{-1}).
\label{likelihood}
\end{equation}
For noise precision $\tau$ in (\ref{likelihood}), we place a Gamma hyperprior over $\tau$, which is
\begin{equation}
    p(\tau) = Ga(\tau|a_0,b_0),
\label{tau}
\end{equation}
where $a_0$ and $b_0$ are set to small values, e.g., $10^{-6}$, so that the Gamma distribution is non-informative. Together with the prior distributions (\ref{U})-(\ref{lambda}), the complete probabilistic model for the matrix completion with dual-graph information is established.

\subsection{Conditional Conjugacy}
In Bayesian inference \cite{beal2003variational,bishop2006pattern,winn2005variational}, closed-form update is viable when the probability distributions of directly connected variables pair in the graphical model is conjugate. While standard Bayesian modeling result \cite{tipping2001sparse,murphy2007conjugate} states that \eqref{lambda} is conjugate to \eqref{U} or \eqref{V}, and \eqref{tau} is conjugate to \eqref{likelihood}, it is not clear whether the newly proposed prior \eqref{U} and \eqref{V} are conjugate to the likelihood function \eqref{likelihood}. This calls for an investigation on the their conjugacy, such that we can determine if inference can be efficiently performed either under variational inference or Gibbs sampling framework \cite{murphy2012machine,gelfand1992bayesian,smith1993bayesian}. When the data matrix $\*M$ is fully observed ($\*\Omega = \*1$) and the Laplacian matrix is identity matrix ($\*L_r =\*L_c=\*I$), the proposed random matrix priors \eqref{U} and \eqref{V} reduce to that of \cite{babacan2012sparse} and meet the conjugacy requirement. However, in our matrix completion task with graph information embedding, both the missing data and non-diagonal Laplacian matrix obscure the conjugacy relationship as stated in the following. 
\begin{property}
The proposed random matrix prior $p(\*U)$ in \eqref{U} is \textbf{not} conjugate to the Gaussian likelihood \eqref{likelihood} conditional on variables $\*\lambda, \*V, \tau$, so as the prior distribution \eqref{V}.
\begin{proof}
The logarithm of the posterior distribution of $\*U$ conditional on $\*M$, $\*V$, $\*\lambda$ and $\tau$ is
    \begin{align}
        &\ln p(\*U|\*M,\*V,\bm{\lambda},\tau)  \propto \ln p(\*M|\*U,\*V,\tau) p(\*U|\bm{\lambda})/p(\*M)\nonumber\\
        =& - \frac{\tau}{2}||\*\Omega\circ (\*M -\*U\*V^T) ||^2_{F} - \frac{1}{2}{\rm tr}(\*\Lambda \*U^T\*L_r\*U) + {\rm const} \nonumber\\
        =& - \frac{\tau}{2} {\rm tr}(\*\Omega\circ(\*U\*V^T \*V \*U^T - 2\*U\*V^T\*M^T)) - \frac{1}{2}{\rm tr}(\*U\*\Lambda \*U^T\*L_r) + {\rm const},
        \label{property_U}
    \end{align}
where $\circ$ denotes Hadamard product, and ${\rm const}$ contains terms not related to $\*U$. As the two terms cannot be merged as one quadratic function of $\*U$ due to the indicator matrix $\*\Omega$ and non-diagonal graph Laplacian matrix $\*L_r$, the proposed random matrix prior is not conjugate to the likelihood function.
\end{proof}
\label{property-noncon1}
\end{property}
\vspace{-0.5cm}
To handle the observation indicator matrix, previous works on matrix completion and tensor completion \cite{babacan2012sparse,zhao2015bayesian,zhao2015bayesianrobust} rely on expressing the likelihood function in terms of rows of $\*U$: $- \frac{1}{2} \sum_i \sum_{j: (i,j)\in \*\Omega} (\*U_{i,:}\*V^T_{j,:}\*V^T_{j,:}\*U_{i,:}^T -2\*U_{i,:}\*V^T_{j,:}\*M_{ij}) + {\rm const}$, which is a quadratic function of rows of $\*U$. If $\*L_r = \*I$, the prior can also be expressed as quadratic function of rows of $\*U$: ${\rm tr}(\*U\*\Lambda \*U^T\*L_r) = \sum_i \*U_{i,:}\*\Lambda\*U^T_{i,:}$. Thus the prior of rows of $\*U$ is conjugate to the likelihood function. Unfortunately, in this work, $\*L_r$ is not an identity matrix.

To get around the effect of non-diagonal $\*L_r$, we notice that the column precision $\*\Lambda$ in \eqref{U} is a diagonal matrix, thus we can express the prior in terms of the columns of $\*U$ as
\begin{equation}
    \ln p(\*U|\*\lambda) = \sum_{i=1}^k \ln p(\*u_i|\lambda_i)\propto  \sum_{i=1}^k -\frac{1}{2}\lambda_i\*u_i^T\*L_r\*u_i,
    \label{property2-U}   
\end{equation}
Now, the question becomes can we express the likelihood function in a quadratic form in terms of the columns of $\*U$.  The following property confirms this possibility.
\begin{property}
The Frobenius norm inside the likelihood function can be expressed as
\begin{align}
    ||\*\Omega\circ (\*M-\*U\*V^T)||^2_F =\*u_i^T {\rm diag}(\*\Omega (\*v_i\circ \*v_i ))\*u_i
    -2 \*u_i^T (\*\Omega\circ(\*M-\sum_{r\ne i}\*u_r\*v_r^T ))\*v_i+ c_{\neg\*u_i},
\label{theorem1-U}
\end{align}
where $c_{\neg\*u_i}$ contains terms not related to $\*u_i$.
\begin{proof}
Assume $\*M_i = \*M-\sum_{r\ne i} \*u_r\*v_r^T$. Let $[\*M_i]_{s,k}$ and $\*\Omega_{s,k}$ be the $(s,k)$ element of $\*M_i$ and $\*\Omega$, respectively. Furthermore, let $u_{s,i}$ and $v_{k,i}$ be the $s^{th}$ element of $\*u_i$ and the $k^{th}$ element of $\*v_i$, respectively. Then, 
\begin{align}
    &||\*\Omega\circ (\*M-\*U\*V^T)||^2_F \nonumber\\
    = \ &||\*\Omega\circ (\*M_i -\*u_i\*v_i^T)||^2_F\nonumber\\
    = \ &\sum_{s=1}^m\sum_{k=1}^n \*\Omega_{s,k}u_{s,i}^2 v_{k,i}^2 -2\sum_{s=1}^m\sum_{k=1}^n \*\Omega_{s,k} [\*M_i]_{s,k} u_{s,i} v_{k,i} + c_{\neg\*u_i}\nonumber\\
    = \ &\sum_{s=1}^m u_{s,i} (\sum_{k=1}^n \*\Omega_{s,k} \times v_{k,i}^2) u_{s,i} -2\sum_{s=1}^m\sum_{k=1}^n u_{s,i}(\*\Omega_{s,k}\times [\*M_i]_{s,k})v_{k,i} + c_{\neg\*u_i}\nonumber\\
    = \ & \*u_i^T
    \begin{bmatrix}
        \sum_{k=1}^n \*\Omega_{1,k}v_{k,i}^2 && \text{0}\\
        & \ddots & \\
        \text{0} & & \sum_{k=1}^n \*\Omega_{m,k}v_{k,i}^2
    \end{bmatrix}
    \*u_i -2 \*u_i^T (\*\Omega\circ\*M_i)\*v_i + c_{\neg\*u_i}\nonumber\\
    =\ &\*u_i^T {\rm diag}\Bigg(
    \begin{bmatrix}
        \*\Omega_{1,1} & \dots & \*\Omega_{1,n}\\
        \vdots& \ddots & \vdots\\
        \*\Omega_{m,1} & \dots& \*\Omega_{m,n}
    \end{bmatrix}
    \begin{bmatrix}
        v_{1,i}^2\\
        \vdots\\
         v_{n,i}^2
    \end{bmatrix}  
    \Bigg)\*u_i -2\*u_i^T (\*\Omega\circ(\*M-\sum_{r\ne i}\*u_r\*v_r^T ))\*v_i + c_{\neg\*u_i}\nonumber\\
    =\ & \*u_i^T {\rm diag}(\*\Omega (\*v_i\circ \*v_i ))\*u_i -2\*u_i^T (\*\Omega\circ(\*M-\sum_{r\ne i}\*u_r\*v_r^T ))\*v_i + c_{\neg\*u_i},
    \label{theorem1-proof1}
\end{align}
where $c_{\neg\*u_i}$ contains terms not related to $\*u_i$.
\end{proof}
\label{Theorem1}
\end{property}
Property \ref{Theorem1} shows the intriguing quadratic form of the likelihood function \eqref{likelihood} with respect to each column $\*u_i$ after fixing other variables, which has a standing alone value for the development of (inexact) block coordinate descent methods \cite{liu2012sparse,xu2013block}, including the VI algorithm in the following section. More importantly, since both the prior of $\*U$ and the likelihood function can both be expressed as quadratic functions with respect to columns of $\*U$, the conditional-conjugacy property of the proposed prior is stated in the following.
\begin{property}
    The proposed prior distributions on $\*u_i$ in \eqref{U} are conjugate to the Gaussian likelihood \eqref{likelihood} conditional on variables $\{\*u_{r \neq i}\}, \*V, \*\lambda, \tau$.
\label{property-cond-conj}
\end{property}
Since $\*U$ and $\*V$ are in the same structure, and a similar result to Property \ref{Theorem1} holds for columns of $\*V$, a similar result to Property \ref{property-cond-conj} can also be derived for $\*v_i$.

\emph{Remark}: Notice that another form of the likelihood function and prior that results in conjugacy is expressing them in terms of individual elements of $\*U$ \cite{lim2007variational}. While this form also leads to closed-form expression of VI or Gibbs sampler, we do not consider this option, since the independence assumption imposed on elements $\{\*U_{ij}\}$ disregards the correlations among them, thus lowering the accuracies of inference. In Bayesian inference, to address the trade-off between the efficiency (tractability) and the accuracy, it is desirable to group as many variables as possible while maintaining (conditional) conjugacy \cite{murphy2012machine}. In the context of dual-graph matrix completion, Property \ref{property-cond-conj} provides such a \textit{good trade-off}.

\section{Inference Algorithm and Insights}
\label{sec3}
Given the newly revealed conditional-conjugacy, in this section, we introduce a column-wise mean-field factorization that differs from previous Bayesian works \cite{babacan2011low,babacan2012sparse,zhao2015bayesian,zhao2015bayesianrobust,li2020bayesian}, and we manage to derive a tailored Bayesian algorithm for the dual-graph and low-rankness embedded matrix completion model with closed-form updates.
\subsection{Mean-field Variational Inference}

The unknown parameter set including factor matrices and hyperparameters is denoted by $\*\Theta = \{\*U,\*V,\bm{\lambda},\tau\}$. Using the likelihood in \eqref{likelihood} and the prior distributions \eqref{U}-\eqref{lambda}, \eqref{tau} the joint distribution can be obtained:
\begin{align}
    p(\*M,\*\Theta) = p(\*M|\*U,\*V,\tau)p(\*U|\bm{\lambda}) p(\*V|\bm{\lambda})p(\bm{\lambda})p(\tau).
\label{MTheta}
\end{align}
The posterior distribution $p(\*\Theta|\*M)$ is then given by:
\begin{equation}
    p(\*\Theta|\*M) = \frac{p(\*M,\*\Theta)}{\int p(\*M,\*\Theta) d\*\Theta}.
\label{theta|M}
\end{equation}

However, the integration in the denominator of (\ref{theta|M}) is analytically intractable and thus prohibits an exact inference. To overcome this problem, variational inference is commonly used to approximate the true posterior distribution with another distribution belonging to a certain family \cite{tzikas2008variational,wainwright2008graphical,zhang2018advances}. To be specific, variational inference seeks a distribution $q(\*\Theta) \in \mathcal{F}$ that is the closest to the true posterior distribution $p(\*\Theta|\*M)$ among a set of distributions in the family $\mathcal{F}$ by minimizing the Kullback-Leibler (KL) divergence:
\begin{align}
    {\rm KL}(q(\*\Theta)||p(\*\Theta|\*M)) &= \int q(\*\Theta) \ln (\frac{q(\*\Theta)}{p(\*\Theta|\*M)})d\*\Theta \nonumber\\
    &= \ln p(\*M) - \mathcal{L},
\label{KL}
\end{align}    
where the model evidence $\ln p(\*M)$ is a constant and the evidence lower bound (ELBO) is defined as $\mathcal{L} = \int q(\*\Theta) \ln (\frac{p(\*M,\*\Theta)}{q(\*\Theta)})d\*\Theta$. Therefore, the KL divergence is minimized when the lower bound is maximized. 
If $\mathcal{F}$ is chosen as the mean-field family $q(\*\Theta)=\prod_j q(\*\Theta_j)$, where $\*\Theta_j \subset \*\Theta$ with $\bigcup \*\Theta_j = \*\Theta$ and $\bigcap \*\Theta_j =\emptyset$, the optimal variational distribution for $\*\Theta_j$ is given by \cite{bishop2006pattern}:
\begin{equation}
    \ln q(\*\Theta_j) = \mathbb{E}_{q(\*\Theta\backslash \*\Theta_j)}[\ln p(\*M,\*\Theta)] + {\rm const}
\label{q(theta)}
\end{equation}
where $\mathbb{E}_{q(\*\Theta\backslash \*\Theta_j)}[\cdot]$ denotes the expectation with respect to the variational distribution over all variables except $\*\Theta_j$. 



\subsection{Column-wise Mean-field Update}
In previous works, the mean-field family $\mathcal{F}$ in the form of $q(\*\Theta) = q(\*U)q(\*V)q(\bm{\lambda})q(\tau)$  is widely employed for matrix completion problems \cite{bishop2006pattern,babacan2012sparse,li2020bayesian}. However, Property \ref{property-noncon1} in previous section prohibits this mean-field assumption in deriving VI algorithm for the proposed probabilistic model. Fortunately, based on the unique conditional conjugacy in Property \ref{property-cond-conj}, we introduce a new mean-field assumption that includes additional factorizations compared with classic mean-field assumption, that is
\begin{align}
    q(\*\Theta) = \prod_{i=1}^k q(\*u_i)\prod_{i=1}^k q(\*v_i)q(\bm{\lambda})q(\tau).
    \label{mean_field_new}
\end{align}
From \eqref{mean_field_new}, the unknown parameter set including latent matrices and hyperparameters is denoted by $\*\Theta = \{\{\*u_i\},\{\*v_i\},\bm{\lambda},\tau\}$. The variational distribution for each $\*\Theta_j$ can be computed using \eqref{q(theta)}. Given the conditional conjugate pair we found in Property \ref{property-cond-conj}, closed-form update for each variable can be obtained for the variational inference. The
detailed derivation of the algorithm is given in the Appendix. Below, we present the key steps in the algorithm, and the
whole algorithm is summarized in Algorithm \ref{alg1}.

\noindent\underline{\textbf{Update of $\*u_i$}:}
\begin{equation}
    q(\*u_i) = \mathcal{N}(\*u_i|\hat{\*\mu}^u_i,\hat{\*\Sigma}^u_i)
\label{Gau-u}
\end{equation}
with $\hat{\*\mu}_i$ and $\hat{\*\Sigma}^u_i$ given by
\begin{align}
    \label{update(U)mu}&\hat{\*\mu}_i^u = \langle \tau\rangle \hat{\*\Sigma}^u_i(\*\Omega\circ(\*M-\sum_{r\ne i} \langle \*u_r\*v_r^T\rangle ))\langle \*v_i\rangle,\\
    \label{update(U)sigma}&\hat{\*\Sigma}^u_i = (\langle\tau\rangle {\rm diag}(\*\Omega\langle \*v_i\circ \*v_i\rangle)+\langle\lambda_i\rangle \*L_r)^{-1},
\end{align}

\noindent\underline{\textbf{Update of $\*v_i$}:} Similar to $\*u_i$, we can find that
\begin{equation}
    q(\*v_i) = \mathcal{N}(\*v_i|\hat{\*\mu}^v_i,\hat{\*\Sigma}^v_i)
\label{Gau-v}
\end{equation}
where
\begin{align}
    \label{update(V)mu}&\hat{\*\mu}^v_i = \langle \tau\rangle \hat{\*\Sigma}^v_i(\*\Omega\circ(\*M-\sum_{r\ne i} \langle \*u_r\*v_r^T\rangle ))^T\langle \*u_i\rangle, \\
    \label{update(V)sigma}&\hat{\*\Sigma}^v_i = (\langle\tau\rangle {\rm diag}(\*\Omega^T\langle \*u_i\circ \*u_i\rangle)+\langle\lambda_i\rangle \*L_c)^{-1}.
\end{align}
\noindent\underline{\textbf{Update of $\*\lambda$}:}
\begin{equation}
    q(\bm{\lambda}) = \prod_{r=1}^k Ga(\lambda_r|\hat{c^r},\hat{d^r})
\label{Ga-lam}
\end{equation}
with $\langle \lambda_i \rangle = \hat{c^r}/\hat{d^r}$
\begin{align}
    \label{update(lam)c}&\hat{c^r} = c_0^r+\frac{m+n}{2},\\
    \label{update(lam)d}&\hat{d^r} = d_0^r+\frac{1}{2}( \langle \*u_r^T\*L_r\*u_r\rangle + \langle \*v_r^T \*L_c \*v_r\rangle).
\end{align}
\noindent\underline{\textbf{Update of $\tau$}:}
\begin{equation}
    q(\tau) = Ga(\tau|\hat{a},\hat{b})
\label{Ga-tau}
\end{equation}
with $\langle \tau \rangle = \hat{a}/\hat{b}$
\begin{align}
    \label{update(tau)a}&\hat{a} = \frac{|\*\Omega|}{2}+a_0,\\
    \label{update(tau)b}&\hat{b} = \frac{1}{2}\langle||\*\Omega\circ (\*M- \*U\*V^T)||^2_F\rangle +b_0,
\end{align}
where $|\*\Omega|$ is the number of observed entries.

\begin{algorithm}[t]
\caption{Variational Inference for the probabilistic graph-embedding matrix completion model} 
\label{alg1}
\begin{algorithmic}
\REQUIRE $\*M$, $\*\Omega$, $\*L_r$, $\*L_c$, $a_0=b_0=10^{-6}$ and $c_0=d_0=10^{-6}$\\
\ENSURE $q(\*u_i)$, $q(\*v_i)$, $q(\tau)$ and $q(\bm{\lambda})$.\\
\STATE Initialize the latent rank $k$ and $\*\Theta$
\WHILE{not converge}
\STATE Update $q(\*u_i)$ via (\ref{update(U)mu})-(\ref{update(U)sigma}) for $i = 1,...,k$;
\STATE Update $q(\*v_i)$ via (\ref{update(V)mu})-(\ref{update(V)sigma}) for $i = 1,...,k$;
\STATE Update $q(\bm{\lambda})$ via (\ref{update(lam)c})-(\ref{update(lam)d});
\STATE Update $q(\tau)$ via (\ref{update(tau)a})-(\ref{update(tau)b});
\STATE Column pruning
\ENDWHILE
\end{algorithmic}
\end{algorithm}
\subsection{Column pruning for complexity reduction} 
In the factorization model, the low-rankness is reflected in the small number of columns in $\*U$ and $\*V$. In the proposed probabilistic model, since there is sparsity in the columns of $\*U$ and $\*V$, we can choose to prune out columns with obviously smaller energy. In particular, if we can compute $||\*u_i||^2+ ||\*v_i||^2$ as the energy of the $i^{th}$ component of factorization, the components with much smaller energies than the maximum component energy can be pruned out. This pruning can be performed at the end of all VI iterations, or performed after each VI iteration. For the latter case, it corresponds to restarting minimization of the KL divergence of a new (but smaller) probabilistic model, with the current variational distributions acting as the initialization of the new minimization. Therefore, the pruning steps will not affect the convergence but also reduces the model size for the algorithm update in the next iteration.
\subsection{Dual-graph embedding helps low-rankness}
 With graph regularization terms $\*u_i^T \*L_r \*u_i$ and $\*v_i^T\*L_c \*v_i$ in (\ref{logu}) being minimized, the hyperparameters $\{\lambda_i\}$, which represents the precision of $i^{th}$ column of $\*U$ and $\*V$,  will likely go to a very large value during inference, since the denominator of computing $\lambda_i$ becomes smaller in (\ref{update(lam)d}). As can be seen from (\ref{update(U)sigma}) and (\ref{update(V)sigma}), when $\lambda_i$ is large enough, covariance matrices $\hat{\*\Sigma}^u_i$ and $\hat{\*\Sigma}^v_i$ both tend to zero. The updating rules (\ref{update(U)mu}) and (\ref{update(V)mu}) further show that in this case, the mean of $\*u_i$ and $\*v_i$ ($\hat{\*\mu}^u_i$ and $\hat{\*\mu}^v_i$) will also be forced to zero. \textit{It is in this way that the dual-graph embedding helps the low-rankness}. 

\subsection{Adaptive dual-graph regularization}
For the remaining high-energy columns of $\*U$ or $\*V$, their corresponding $\{\lambda_i\}$ are not necessarily the same. This provides a trade-off between model fitting (likelihood) and graph regularization (prior) for each column vector. As we can see from \eqref{update(U)sigma} and \eqref{update(V)sigma}, the ratio of noise precision $\tau$ to hyperparameter $\lambda_i$ decides the relative importance of dual-graph information when updating each $\*u_i$ and $\*v_i$. A high $\lambda_i/\tau$ ratio represents a low signal-to-noise ratio (SNR) for this component, and thus the algorithm puts heavier weight on dual-graph information regularization. Reversely, when $\lambda_i/\tau$ ratio is low, which means a high SNR, the algorithm is more confident with the observed data than the graph structures, and therefore the algorithm will rely more on the data than the dual-graph regularization.

This corresponds to an adaptive regularization for each component $\*u_i$ or $\*v_i$ as oppose to the single regularization applied to all components in existing models \cite{rao2015collaborative,strahl2020scalable}. The relative confidence towards observation and dual-graph information guides the completion algorithm to learn more effectively, contributing to an enhanced model robustness and improved performance for the proposed model.

It is worth emphasizing that the balancing between confidence of observation and dual-graph information is automatic, which has not been provided by any algorithm before.


\subsection{Complexity Analysis}
Given a partially observed data matrix $\*M \in \mathbb{R}^{m\times n}$, the computational cost is dominated by the matrix inverse in updates of $q(\*u_i)$ and $q(\*v_i)$, thus the overall time complexity is $\mathcal{O}(k\max (m^3,n^3))$ where $k$ is the initial latent rank. It is worth noting that conducting column pruning at the end of each iteration of the algorithm will reduce the computational complexity at later iterations, especially for low-rank completion tasks.

\section{Experiments}
\label{sec4}
In this section, we present experimental results on both synthetic data and real-world data for various applications. We compare our dual-graph regularized Bayesian matrix completion method (BMCG) with several state-of-the-art methods: 
\begin{itemize}[leftmargin=*]
\item MC on Graphs \cite{kalofolias2014matrix}: an ADMM-based optimization method which makes use of dual-graph information, whose objective function is introduced in equation (\ref{mc05}); 
\item L1MC \cite{shi2017rank} and LMaFit \cite{wen2012solving}: optimization-based matrix completion methods and L1MC is capable of rank selection;
\item BPMF \cite{salakhutdinov2008bayesian} and VBMC \cite{babacan2012sparse}: two classic Bayesian matrix factorization and matrix completion methods;
\item BMC-GP-GAMP \cite{yang2018fast} and KPMF \cite{zhou2012kernelized}: two probabilistic methods using graph information; 
\item GRALS \cite{rao2015collaborative} and GPMF \cite{strahl2020scalable}:  dual-graph regularized alternating EM methods based on (\ref{opt-fac}); GPMF is derived based on GRALS with an additional graph edge pruning step.
\end{itemize}
Unless stated otherwise, the initial rank for the proposed method BMCG, LMaFit, L1MC and VBMC is set to 100, since they can automatically reduce the rank. For BPMF, KPMF, GRALS and GPMF, $k$ is set to 10, which is the default choice reported in existing works \cite{rao2015collaborative,strahl2020scalable}. All the results in subsections \ref{subA} and \ref{subB} are presented in terms of root-mean-square error (RMSE) \cite{hyndman2006another}.

\subsection{Validation on Synthetic Data}
\labelsubseccounter{subA}
\begin{figure*}[htb]
\centering
\subfloat[\label{fig2a}]{\includegraphics[width=0.33\linewidth]{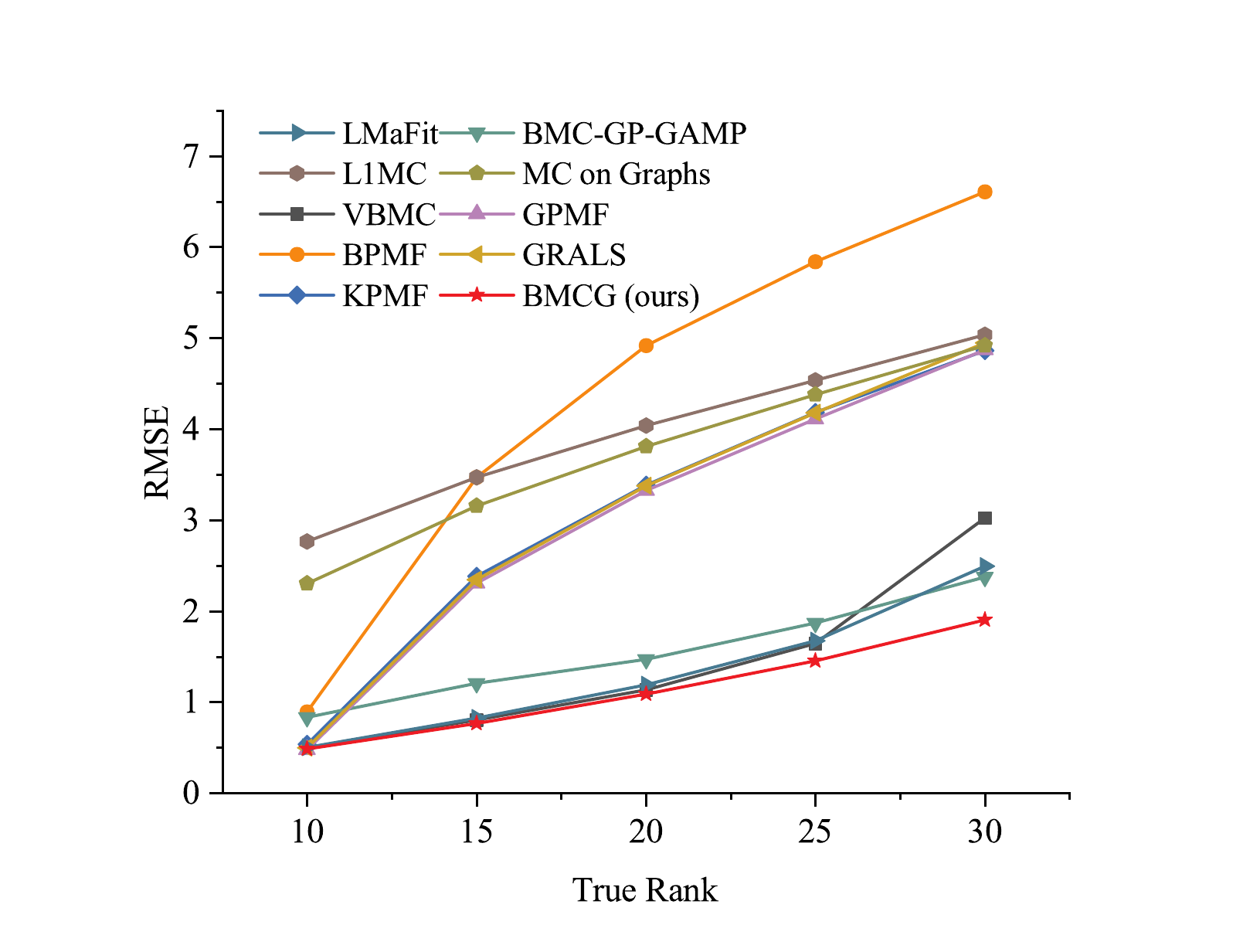}}
\hfill
\subfloat[\label{fig2b}]{\includegraphics[width=0.33\linewidth]{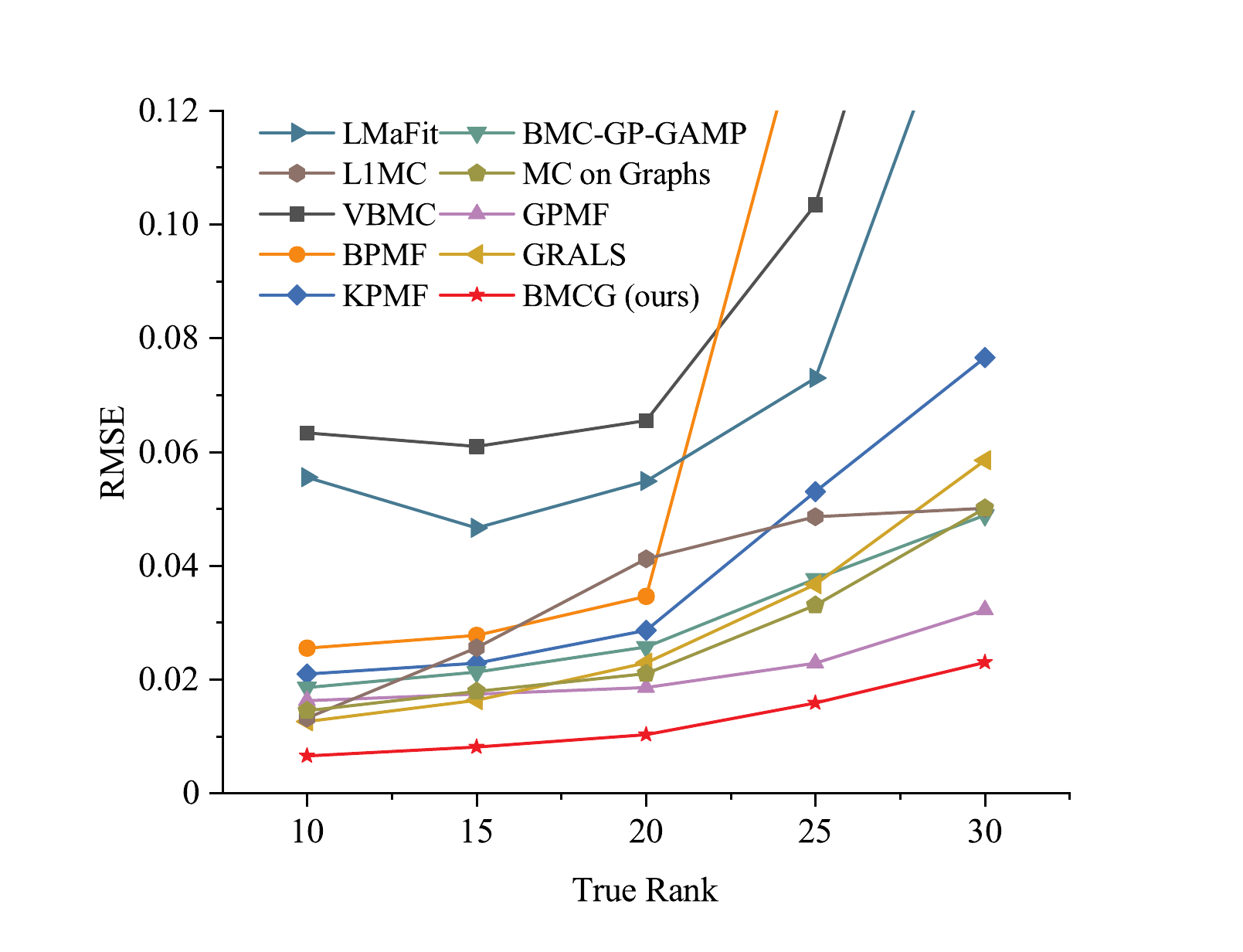}}
\hfill
\subfloat[\label{fig2c}]{\includegraphics[width=0.33\linewidth]{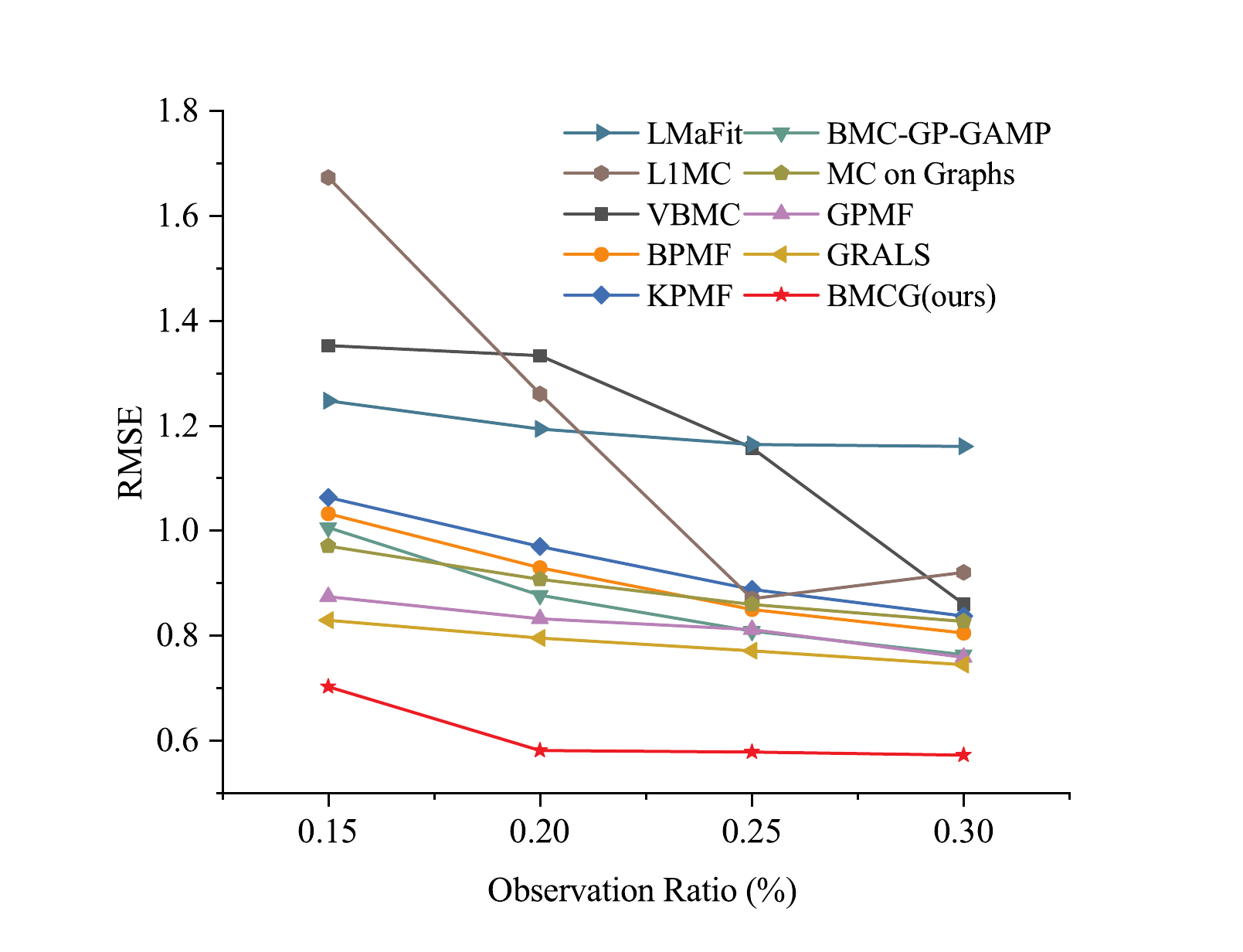}}
\hfill
\centering
\caption{Synthetic data: (a) $500\times500$ random matrix: RMSE vs. the true rank of matrix, observation ratio $= 0.2$, SNR $= 10$ dB; (b) $500\times500$ graph-structured matrix: RMSE vs. the true rank of matrix, observation ratio $ = 0.2$, SNR $= 10$ dB; (c) Netflix-like data: RMSE vs. the observation ratio} 
\label{fig:RMSE}
\end{figure*}
We conduct experiments on 
\begin{enumerate}[leftmargin=*]
\item $500 \times 500$ synthetic matrix, where each element in factor matrices $\*U \in \mathbb{R}^{500\times k}$ and $\*V \in \mathbb{R}^{500\times k}$ is independently drawn from a standard Gaussian distribution $\mathcal{N}(0,1)$. The data is corrupted by a noise matrix $\*E$ whose elements are i.i.d. zero-mean Gaussian distributed with variance $\sigma^2$. The SNR is defined as $10\log_{10}(\frac{\sigma_X^2}{\sigma^2})$ \cite{zhao2015bayesian}, where $\sigma_X^2 = {\rm var}({\rm vec}(\*X))$. The observed entries are sampled randomly and recorded in the indicator matrix $\*\Omega$. We construct 10-nearest neighbor graphs through the function $\rm gsp\_nn\_graph(\cdot)$ provided in Graph Signal Processing Toolbox \cite{perraudin2014gspbox}.
\item $500 \times 500$ graph-structured matrix, where $\*U$ and $\*V$ are generated based on equation (\ref{U})-(\ref{V}) with $\*\Lambda=\*I$ and a constant graph Laplacian matrice \cite{yang2018fast,kheradmand2014general}
\begin{align}
    \label{Asyn}\*A_{ij} &= {\rm exp}(-\frac{|i-j|^2}{\theta^2})\\
    \label{Lsyn}\*L &= \*D - \*A + \hat{\epsilon}\*I
\end{align}
where $\*D$ is the diagonal degree matrix with $\*D_{ii} = \sum_j \*A_{ij}$, parameters $\hat{\epsilon}$ and $\theta$ are set to $10^{-6}$ and $\sqrt{3}$ respectively. The matrix is corrupted by noise $\*E$ with i.i.d. Gaussian elements with zero mean and variance $\sigma^2$.
\item $100 \times 200$ Netflix-like rating data matrix \cite{kalofolias2014matrix}  with integer entries ranging from 1 to 5, representing the pairwise ratings between users and movies. The column and row graph information are provided with the dataset.
\end{enumerate}

As shown in Fig. \ref{fig:RMSE}\subref{fig2a}, when the true ranks are unavailable and data have only weak graph structures (since the data are i.i.d.), the algorithms with the ability to induce low-rankness outperform those with pre-defined ranks. Low-rankness is a way of controlling the complexity of the model and is important in missing data imputation. Among the algorithms with automatic rank reduction, the proposed algorithm (red line) achieves the best performance, since it effectively exploits the dual-graph information, even in weak graphs. 

In comparison, in Fig. \ref{fig:RMSE}\subref{fig2b}, if data have informative graph structures (since data are generated using \eqref{U}-\eqref{V}), it can be observed that incorporating graph information significantly improves the accuracies of missing data recovery. Again, the proposed algorithm exhibits the best performance, with at least 28\% improvement compared to the second best accuracy obtained from GRALS or GPMF. The results in Fig. \ref{fig:RMSE}\subref{fig2a} and \ref{fig:RMSE}\subref{fig2b} corroborate the analysis of the proposed new prior on the simultaneous embedding of low-rankness and dual-graph structure, and the performance of the proposed algorithm does not deteriorate in data having only weak graph structure. Moreover, the superior performance of the proposed method is observed in a wide range of true ranks.


In Fig. \ref{fig:RMSE}\subref{fig2c}, we present completion results on the synthetic Netflix-like dataset. As shown in the figure, the proposed method achieves the most accurate rating prediction. Besides, the algorithms that perform closest to the proposed algorithm are dual-graph regularized (GRALS, GPMF), and then followed by single graph regularized method (BMC-GP-GAMP). In contrast, the performances of algorithms that only induce low-rankness (VBMC, LMaFit), which achieves second and third best in Fig. \ref{fig:RMSE}\subref{fig2a} for i.i.d. data, degrade to almost the worst ones in Fig. \ref{fig:RMSE}\subref{fig2c} due to their lack of graph embedding ability.

\subsection{Recommendation Systems}
\labelsubseccounter{subB}
\begin{table}[tbp]
\centering
\caption{Result summary on recommendation system datasets (RMSE). Bold = best result}
\label{tab:recom}
\resizebox{0.7\linewidth}{!}{%
\begin{tabular}{cccccc}
  \hline
 &
  \begin{tabular}[c]{@{}c@{}}FLIXSTER\\ (3K$\times$3K)\end{tabular} &
  \begin{tabular}[c]{@{}c@{}}DOUBAN\\ (3K$\times$3K)\end{tabular} &
  \begin{tabular}[c]{@{}c@{}}MOVIELENS 100K\\ (943$\times$1682)\end{tabular} &
  \begin{tabular}[c]{@{}c@{}}MOVIELENS 1M\\ (6K$\times$4K)\end{tabular} \\ \hline
BMCG (ours)  & \textbf{0.8748} & 0.7366                           & \textbf{0.9044} & \textbf{0.8442} \\
MC on Graphs & 1.0371 & 0.9011                           & 1.0869                           & 1.0460                           \\
L1MC         & 2.3334 & 1.5867                           & 2.6197                           & 1.6400                           \\
LMaFit       & 1.0927 & 0.7671                           & 0.9682                           & 0.9215                           \\
GRALS       & 0.9152 & 0.7504                           & 0.9211                           & 0.8588                           \\
GPMF         & 0.8857 & 0.7497                           & 0.9336                           & 0.8657          \\
BMC-GP-GAMP  & 3.3250 & 2.7076                           & 1.0104                           & 0.9527                           \\
KPMF         & 0.9218 & \textbf{0.7328} & 0.9606                           & 0.8618                           \\
BPMF        & 1.0866 & 0.8404 & 0.9469 & 0.8754 \\
VBMC         & 1.0789 & 0.7372                           &  0.9295                           &  0.9295                           \\ \hline                     
\end{tabular}%
}
\end{table}
In this section we analyze four famous datasets in the field of recommendation systems: 3k by 3k sub-matrices of Flixster \cite{jamali2010matrix} and Douban \cite{ma2011recommender,monti2017geometric}; MovieLens $100k$ and $1M$ \cite{harper2015movielens}. Following \cite{rao2015collaborative}, for the two MovieLens datasets, we construct 10-nearest neighbor user and movie graphs based on user features (age, gender, and occupation) and item features (genres) respectively. For Flixster, the two 10-nearest neighbor graphs are bulit from the ratings of the original matrix. On the other hand, for the Douban dataset, we only use the sparse user graph from the provided social network.

Table \ref{tab:recom} summarizes the performance of different methods on recommendation tasks. It can be seen that the proposed method BMCG has the best performances in three of the datasets, and is the second best for Douban dataset. For GRALS and GPMF, we used the recommended settings in \cite{strahl2020scalable}. Since the hyper-parameters of GRALS and GPMF are heavily tuned and tailored for the recommendation system datasets, their performances are close to those of the proposed algorithm. However, the proposed algorithm does not require any tunning, and obtain the results in Table \ref{tab:recom} with only one execution.
\subsection{Genotype Imputation}

\begin{table}[t]
\centering
\caption{Result summary on genotype imputation under different observed ratio (error rate). Bold = best result}
\label{tab:gene}
\resizebox{0.4\linewidth}{!}{%
\begin{tabular}{ccccc}
\hline
             & 10\%       & 20\%       & 50\%       \\ \hline
BMCG (ours)  & \textbf{0.0373} & \textbf{0.0258} & \textbf{0.0174} \\
MC on Graphs & 0.1567  & 0.1048  & 0.0517 \\
L1MC         & 0.2517   & 0.1311  & 0.0391 \\
LMaFit       & 0.2517  & 0.1498  & 0.2462  \\
GRALS       & 0.5140  & 0.6324  & 0.7200  \\
GPMF       & 0.7483  & 0.7479  & 0.7490  \\
BMC-GP-GAMP  & 0.1516  & 0.0574 & 0.0216 \\
KPMF         & 0.1600  & 0.1087  & 0.0439 \\
BPMF         & 0.5241  & 0.6659  & 0.7370 \\
VBMC         & 0.1250  & 0.0588  & 0.0226 \\\hline
\end{tabular}%
}
\end{table}
\begin{table*}[t]
\centering
\caption{Image Inpainting (PSNR/SSIM). Bold = best performance}
\label{tab:imgrec}
\resizebox{\textwidth}{!}{%
\begin{tabular}{cccccccccc}
\hline
             & house          & peppers          & baboon         & barbara        & bridge         & facade         & airplane       & church         & wall           \\ \hline
BMCG (ours)  & \textbf{20.7708/0.9992} & \textbf{20.5071/0.9991}   & \textbf{19.5435/0.9991} & \textbf{20.2072/0.9993} & \textbf{19.4531/0.9989} & \textbf{20.1365/0.9994} & \textbf{20.1453/0.9992} & 19.5991/\textbf{0.9984} & \textbf{18.7717/0.9988} \\
MC on Graphs & 4.3750/0.9546  & 6.6271/0.9688    & 5.4798/0.9629  & 6.3946/0.9708  & 9.3438/0.9835  & 5.9108/0.9636  & 1.8839/0.9266  & 4.2704/0.9453  & 6.8264/0.9697  \\
L1MC         & 19.5110/0.9981 & 12.6088/0.9844 & 14.6659/0.9888 & 7.8225/0.9729  & 10.5303/0.9830 & 11.2043/0.9793 & 17.9984/0.9978 & 18.1393/0.9965 & 14.1414/0.9924 \\
LMaFit       & 7.8145/0.9988  & 7.8713/0.9680    & 8.2523/0.9649  & 7.8220/0.9720  & 9.3913/0.9826  & 7.4176/0.9685  & 3.7020/0.9387  & 6.1006/0.9472  & 7.3909/0.9712  \\
GRALS        & 18.1822/0.9988 & 17.3200/0.9984   & 17.8022/0.9988 & 18.1858/0.9991 & 17.5501/0.9987 & 19.8915/0.9990 & 19.4593/0.9990 & 17.7695/0.9982 & 17.1057/0.9983 \\
GPMF         & 17.3920/0.9979 & 13.3687/0.9947   & 15.0444/0.9969 & 14.5834/0.9964 & 15.4525/0.9969 & 18.8089/0.9989 & 15.8009/0.9974 & 15.3881/0.9962 & 15.2186/0.9971 \\
BMC-GP-GAMP  & 20.0749/0.9988 & 18.4687/0.9981   & 19.1917/0.9989 & 18.2256/0.9985 & 19.1906/0.9986 & 19.4272/0.9992 & 19.0971/0.9986 & \textbf{20.4175/0.9984} & 17.0213/0.9978 \\
KPMF         & 17.4634/0.9981 & 14.5266/0.9963   & 16.0051/0.9978 & 15.4919/0.9973 & 15.8074/0.9973 & 20.0450/0.9992 & 16.6568/0.9981 & 15.5014/0.9968 & 15.6994/0.9976 \\
BPMF         & 13.7860/0.9932 & 13.3002/0.9918   & 13.8773/0.9927 & 13.3284/0.9927 & 13.9810/0.9933 & 13.8794/0.9945 & 13.4697/0.9932 & 14.5168/0.9920 & 14.1474/0.9935 \\
VBMC         & 17.4998/0.9974 & 13.0452/0.9944   & 15.2448/0.9971 & 14.3214/0.9963 & 14.6432/0.9961 & 18.3471/0.9987 & 15.3214/0.9972 & 15.6427/0.9957 & 14.6805/0.9967 \\ \hline
\end{tabular}%
}
\end{table*}
\begin{figure*}[t]
\centering
\includegraphics[width=0.9\linewidth]{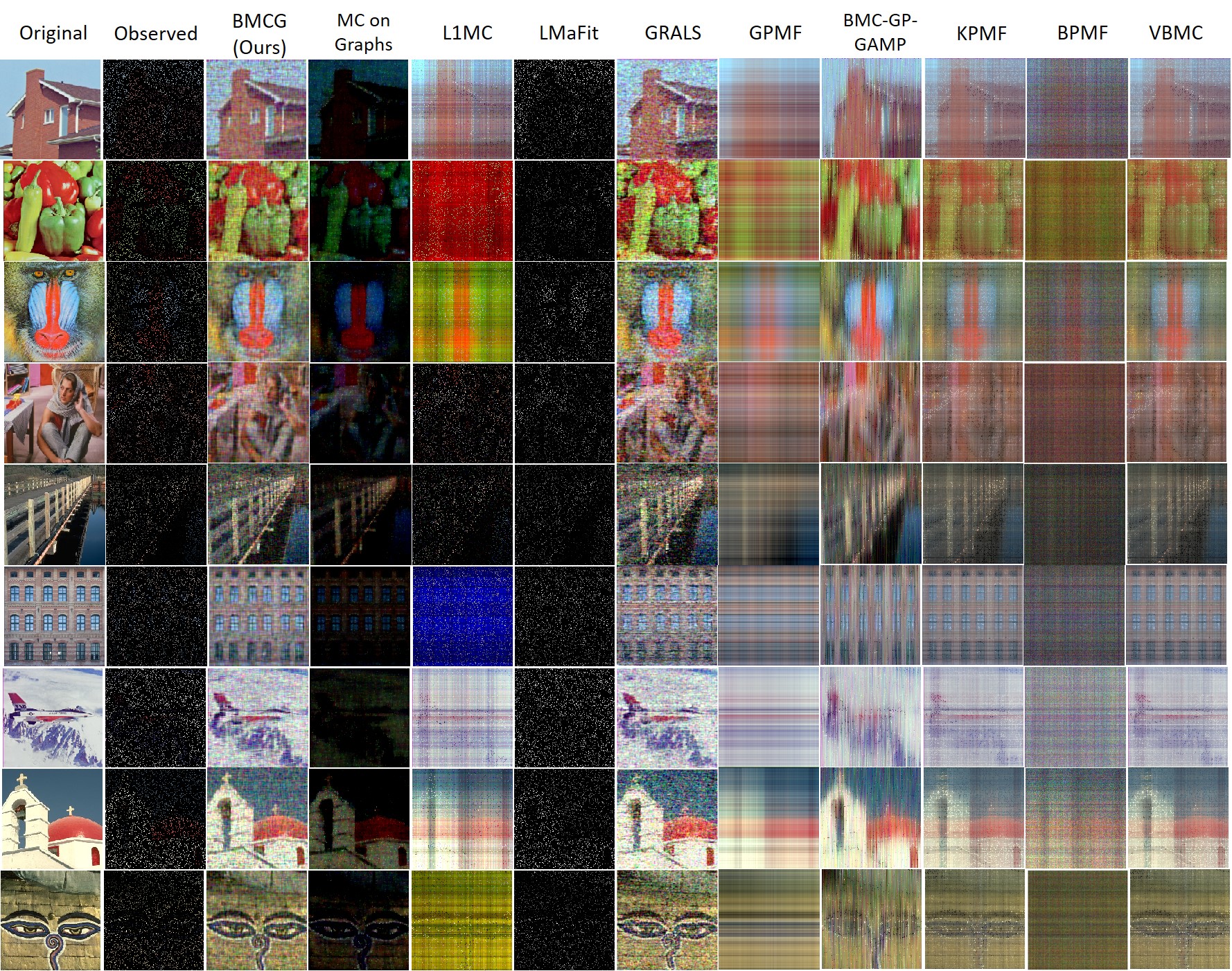}
\caption{The image inpaiting results with noise SNR $= 5$ dB and 10\% pixels observed}
\label{fig:imgrec}
\end{figure*} 
In this section, the application of genotype imputation based on gene data is used to compare the performance of different methods. The dataset Chr22 \cite{jiang2016sparrec} is a 790$\times$112 matrix which contains haplotype information. Inputing missing genetic data shares many features with application of recommendation systems as they both take advantage of the low-rank and/or low number of co-clusters structures of datasets. \cite{chi2013genotype}. We evaluate the imputing performance by imputation error rate defined as:
\begin{equation}
    {\rm Error\ Rate} = \frac{N_{{\rm imputed}\neq {\rm true}}}{N_{{\rm unobserved}}}
\end{equation}
where $N_{{\rm imputed}\neq {\rm true}}$ denotes number of wrong imputations after compared with ground truth, and $N_{{\rm unobserved}}$ denotes the number of unobserved entries. We only consider a 10-nearest neighbor row graph in this experiment, which is constructed from the labels of the original matrix. We consider different cases where 10\%, 20\%, and 50\% entries are observed. The imputing result is reported in Table \ref{tab:gene}, from which it is clear that the proposed method gives the best performances under every observed ratio. Also, it can be noticed that while  GRALS and GPMF perform well in recommendation systems, they perform almost the worst in Gene dataset. This demonstrates the importance of parameter tuning in these optimization based methods. In contrast, the proposed method automatically adapts to the new dataset and gives the best performance without any tuning.

\subsection{Image Inpainting}
Finally, we examine the performance of different methods on image inpainting tasks. We conduct the experiments on the benchmark images and images from Berkeley segmentation dataset \cite{MartinFTM01}. For each image, we add noise with SNR $=5$ dB, and randomly choose 10\% of of pixels as observations. We formulate the underlying dual-graph information of images as \eqref{Asyn}-\eqref{Lsyn}.
The peak signal to noise ratio (PSNR) and the structural similarity (SSIM) index are shown in Table \ref{tab:imgrec}. The recovered images are shown in Fig. \ref{fig:imgrec}.

By visual inspection of the recovered images in Fig. \ref{fig:imgrec}, it is obvious that only the proposed method and GRALS recover recognizable images. However, Table \ref{tab:imgrec} shows that the proposed algorithm achieves an average of 1.76dB, up to 3.2dB higher PSNR than the GRALS.
\section{Conclusion}
\label{sec_con}
BMCG, a probabilistic matrix completion model that simultaneously embeds the dual-graph information and low-rankness, has been proposed in this paper. Theoretical analysis has been presented to show that the dual-graph information and low-rank property are effectively conveyed in the proposed prior and its advantages over other priors have been justified. With a detailed investigation of conditional conjugacy between the newly designed prior and likelihood function, an efficient algorithm has been derived using variational inference that achieves automatic rank reduction and is free from regularization parameters tuning. The update equations reveal the insights of low-rankness and graph structure exploitation. Extensive experimental results on various data analytic tasks have showcased the superior missing data imputation accuracy compared to other state-of-the-art methods.

\bibliographystyle{IEEEtran}
\bibliography{main}
\appendix[Derivation of the Column-wise Variational Inference]
The unknown parameter set including latent matrices and hyperparameters is denoted by $\*\Theta = \{\{\*u_i\},\{\*v_i\},\bm{\lambda},\tau\}$.
The variational distribution for each $\*\Theta_j$ is given by:
\begin{equation}
    \ln q(\*\Theta_j) = \mathbb{E}_{q(\*\Theta\backslash \*\Theta_j)}[\ln p(\*M,\*\Theta)] + {\rm const}
\label{q(theta)_up}
\end{equation}
where $\mathbb{E}_{q(\*\Theta\backslash \*\Theta_j)}[\cdot]$ denotes the expectation w.r.t the $q$ distributions over all variables except $\*\Theta_j$. The logarithm of the joint probability density function $\ln p(\*M,\*\Theta)$ is
\begin{align}
&\ln p(\*M,\{\{\*u_i\},\{\*v_i\},\bm{\lambda},\tau)  \nonumber\\
=& \ln p(\*M|\{\{\*u_i\},\{\*v_i\},\tau) \sum_{r=1}^k p(\*u_r|\lambda_r)\sum_{r=1}^k p(\*v_r|\lambda_r)p(\tau) p(\bm{\lambda})\nonumber\\
\propto& (\frac{|\*\Omega|}{2} +a_0-1) \ln \tau
 - \frac{\tau}{2}||\*\Omega\circ(\*M -\sum_{r=1}^k\*u_r\*v_r^T) ||^2_{F}\nonumber\\
&-\sum_{r=1}^k \frac{1}{2} \lambda_r \*u_r^T\*L_r\*u_r- \sum_{r=1}^k\frac{1}{2} \lambda_r \*v_r^T\*L_c\*v_r \nonumber\\
&+ \sum_{r=1}^k [(\frac{m+n}{2}+c_0^r-1)\ln \lambda_r]-\sum_{r=1}^k d_0^r\lambda_r -b_0\tau + {\rm const}.
\label{logjoint_up2}
\end{align}
By substituting (\ref{logjoint_up2}) into (\ref{q(theta)_up}), we obtain $q(\*u_i)$:
\begin{align}
    \ln q(\*u_i) =&  \mathbb{E}_{q(\*\Theta \setminus \*u_i)}\{-\frac{\tau}{2}||\*\Omega\circ (\*M-\sum_{r=1}^{k} \*u_r\*v_r^T)||^2_F-\frac{1}{2}\sum_{r=1}^{k}\lambda_r\*u_r^T\*L_r\*u_r \} \nonumber \\
    =&\mathbb{E}_{q(\*\Theta \setminus \*u_i)} \{-\frac{\tau}{2}||\*\Omega\circ (\*M-\sum_{r\ne i} \*u_r\*v_r^T-\*u_i\*v_i^T)||^2_F -\frac{1}{2}\lambda_i\*u_i^T\*L_r\*u_i\}\nonumber\\
    =&-\frac{1}{2}\*u_i^T[\langle\tau\rangle {\rm diag}(\*\Omega\langle \*v_i\circ \*v_i\rangle)+\langle\lambda_i\rangle \*L_r]\*u_i+\*u_i^T\langle \tau\rangle (\*\Omega\circ(\*M-\sum_{r\ne i} \langle \*u_r\*v_r^T\rangle ))\langle \*v_i\rangle 
\label{q(u)_up1}
\end{align}
where $\langle \cdot \rangle$ represents expectation and $\circ$ denotes Hadamard product. We can see from (\ref{q(u)_up1}) that each column $\*u_i$ follows a Gaussian distribution, which is
\begin{equation}
    q(\*u_i) = \mathcal{N}(\*u_i|\hat{\*\mu}^u_i,\hat{\*\Sigma}^u_i)
\label{Gau-u_up}
\end{equation}
with $\hat{\*\mu}_i$ and $\hat{\*\Sigma}^u_i$ given as
\begin{align}
    &\hat{\*\mu}_i^u = \langle \tau\rangle \hat{\*\Sigma}^u_i(\*\Omega\circ(\*M-\sum_{r\ne i} \langle \*u_r\*v_r^T\rangle ))\langle \*v_i\rangle,\\
    &\hat{\*\Sigma}^u_i = (\langle\tau\rangle {\rm diag}(\*\Omega\langle \*v_i\circ \*v_i\rangle)+\langle\lambda_i\rangle \*L_r)^{-1}.
\label{update(U)_up}
\end{align}
Similarly, we can prove that the columns of factor matrix $\*V$, i.e. $\*v_i$, follows a Gaussian distribution
\begin{equation}
    q(\*v_i) = \mathcal{N}(\*v_i|\hat{\*\mu}^v_i,\hat{\*\Sigma}^v_i)
\label{Gau-v_up}
\end{equation}
where
\begin{align}
    &\hat{\*\mu}^v_i = \langle \tau\rangle \hat{\*\Sigma}^v_i(\*\Omega\circ(\*M-\sum_{r\ne i} \langle \*u_r\*v_r^T\rangle ))^T\langle \*u_i\rangle, \\
    &\hat{\*\Sigma}^v_i = (\langle\tau\rangle {\rm diag}(\*\Omega^T\langle \*u_i\circ \*u_i\rangle)+\langle\lambda_i\rangle \*L_c)^{-1}.
\label{update(V)_up}
\end{align}
The expression of $q(\bm{\lambda})$ can be found as
\begin{align}
\ln q(\bm{\lambda}) = 
&\mathbb{E}_{q(\*\Theta\setminus\bm{\lambda})}[\ln p(\*M,\{\{\*u_i\},\{\*v_i\},\bm{\lambda},\tau)]+{\rm const} \nonumber\\
= &\mathbb{E}_{q(\*\Theta\setminus\bm{\lambda})}\{-\frac{1}{2}\sum_{r=1}^k \lambda_r \*u_r^T\*L_r\*u_r-\frac{1}{2}\sum_{r=1}^k \lambda_r \*v_r^T\*L_c\*v_r \nonumber\\
&+\sum _{r=1}^k[(\frac{m+n}{2}+c_0^r-1 )\ln \lambda _r]-\sum _{r=1}^kd_0^r\lambda_r\}\nonumber\\
=&\mathbb{E}\sum _{r=1}^k\{(\frac{m+n}{2}+c_0^r-1 )\ln \lambda _r-[d_0^r+\frac{1}{2}( \*u_r^T\*L_r\*u_r+ \*v_r^T\*L_c\*v_r)]\lambda_r\}\nonumber\\
= &\sum_{r=1}^k\{(c^r+\frac{m+n}{2}-1)\ln \lambda_r-[d^r+\frac{1}{2}( \langle \*u_r^T\*L_r\*u_r\rangle + \langle \*v_r^T\*L_c\*v_r\rangle)]\lambda_r\}.
\label{q(lam)_up}
\end{align}
Note that the expectation of the quadratic terms are calculated as $\langle \*u_r^T\*L_r\*u_r\rangle = tr[(\langle\*\Sigma^u_r\rangle + \langle \*u_r\rangle\langle \*u_r^T\rangle)\*L_r]$ and $\langle \*v_r^T \*L_c \*v_r\rangle = tr[(\langle\*\Sigma^v_r\rangle + \langle \*v_r\rangle\langle \*v_r^T\rangle)\*L_c]$.
Finally, we can easily have the update rules for $\tau$ from
\begin{align}
    \ln q(\tau) = &(\frac{|\*\Omega|}{2}+a-1 )\ln \tau -(\frac{1}{2}\langle||\*\Omega\circ (\*M- \*U\*V^T)||^2_F\rangle +b)\tau
\label{q(tau)_up}
\end{align}
which obviously shows that $\tau$ follows a Gamma distribution
\begin{equation}
    q(\tau) = Ga(\tau|\hat{a},\hat{b})
\label{Ga-tau_up}
\end{equation}
where
\begin{align}
    &\hat{a} = \frac{|\*\Omega|}{2}+a,\\
    &\hat{b} = \frac{1}{2}\langle||\*\Omega\circ (\*M- \*U\*V^T)||^2_F\rangle +b.
    \label{update(tau)_up}
\end{align}

\end{document}